\documentclass[twocolumn,natbib]{svjour3}          
\smartqed  

\usepackage{graphicx}
\usepackage{enumitem}
\usepackage{comment}
\usepackage{amsmath,amssymb} 
\usepackage{mathtools, cuted}
\usepackage{microtype}
\usepackage{multirow}
\usepackage{bm}
\usepackage{algorithm}
\usepackage{listings}
\usepackage{tikz}
\usepackage{pgfplots}
\usetikzlibrary{intersections}
\usepgfplotslibrary{fillbetween}
\pgfplotsset{compat=1.12}
\tikzset{venn circle/.style={draw,circle,minimum width=3cm,draw=none,fill=#1!40,name path=#1}}

\usepackage{xcolor}
\definecolor{mydarkblue}{rgb}{0,0.08,0.45}
\definecolor{myblue}{RGB}{103,169,207}
\definecolor{myred}{RGB}{239,138,98}

\usepackage{hyperref}
\hypersetup{%
colorlinks=true,
linkcolor=mydarkblue,
citecolor=mydarkblue,
filecolor=mydarkblue,
urlcolor=mydarkblue}
\usepackage{multirow}
\usepackage{float}
\usepackage{blindtext}
\usepackage{algorithm}
\usepackage{algorithmic}
\usepackage{xcolor}
\DeclareMathOperator*{\argmax}{arg\,max}

\usepackage{color}
\definecolor{mypink}{RGB}{251, 180, 174}
\definecolor{mygreen}{RGB}{77, 175, 74}
\definecolor{myblue}{RGB}{55, 126, 184}

\newcommand{\eatme}[1]{ }

\newlength\myindent
\usepackage{xurl}

\begin{document}

\title{Learning Scene Dynamics from Point Cloud Sequences  
}


\author{Pan He\thanks{Modern Artificial Intelligence and Learning Technologies Lab (MALT-Lab), 	Department of Computer \& Information Science \& Engineering, University of Florida, Gainesville, FL, USA}        \and
        Patrick Emami \and
        Sanjay Ranka \and
        Anand Rangarajan
}


\institute{Pan He \at
\email{pan.he@ufl.edu}           
\and
Patrick Emami \at
\email{pemami@ufl.edu} 
\and
Sanjay Ranka \at 
\email{ranka@cise.ufl.edu} 
\and
Anand Rangarajan \at 
\email{anand@cise.ufl.edu} 
\and
}

\date{Received: date / Accepted: date}

\maketitle
\begin{abstract}

Understanding 3D scenes is a critical prerequisite for autonomous agents. Recently, LiDAR and other sensors have made large amounts of data available in the form of temporal sequences of point cloud frames. In this work, we propose a novel problem---\emph{sequential} scene flow estimation (SSFE)---that aims to predict 3D scene flow for all pairs of point clouds in a given sequence. This is unlike the previously studied problem of scene flow estimation which focuses on two frames.

We introduce the SPCM-Net architecture, which solves this problem by computing multi-scale spatiotemporal correlations between neighboring point clouds and then aggregating the correlation across time with an order-invariant recurrent unit. Our experimental evaluation confirms that recurrent processing of point cloud sequences results in significantly better SSFE compared to using only two frames. Additionally, we demonstrate that this approach can be effectively modified for sequential point cloud forecasting (SPF), a related problem that demands forecasting future point cloud frames. 

Our experimental results are evaluated using a new benchmark for both SSFE and SPF consisting of synthetic and real datasets. Previously, datasets for scene flow estimation have been limited to two frames. We provide non-trivial extensions to these datasets for multi-frame estimation and prediction. Due to the difficulty of obtaining ground truth motion for real-world datasets, we use self-supervised training and evaluation metrics. We believe that this benchmark will be pivotal to future research in this area. All code for benchmark and models will be made accessible at (\url{https://github.com/BestSonny/SPCM}).

\keywords{3D Deep Learning \and Scene Dynamics \and Point Cloud Processing \and Scene Flow Estimation \and Spatiotemporal Learning \and Self-Supervised Learning}
\end{abstract}

\section{Introduction}\label{sec:introduction}

\begin{figure*}
    \centering
    \includegraphics[clip, trim=0cm 0.0cm 0cm 0.0cm, width=\textwidth]{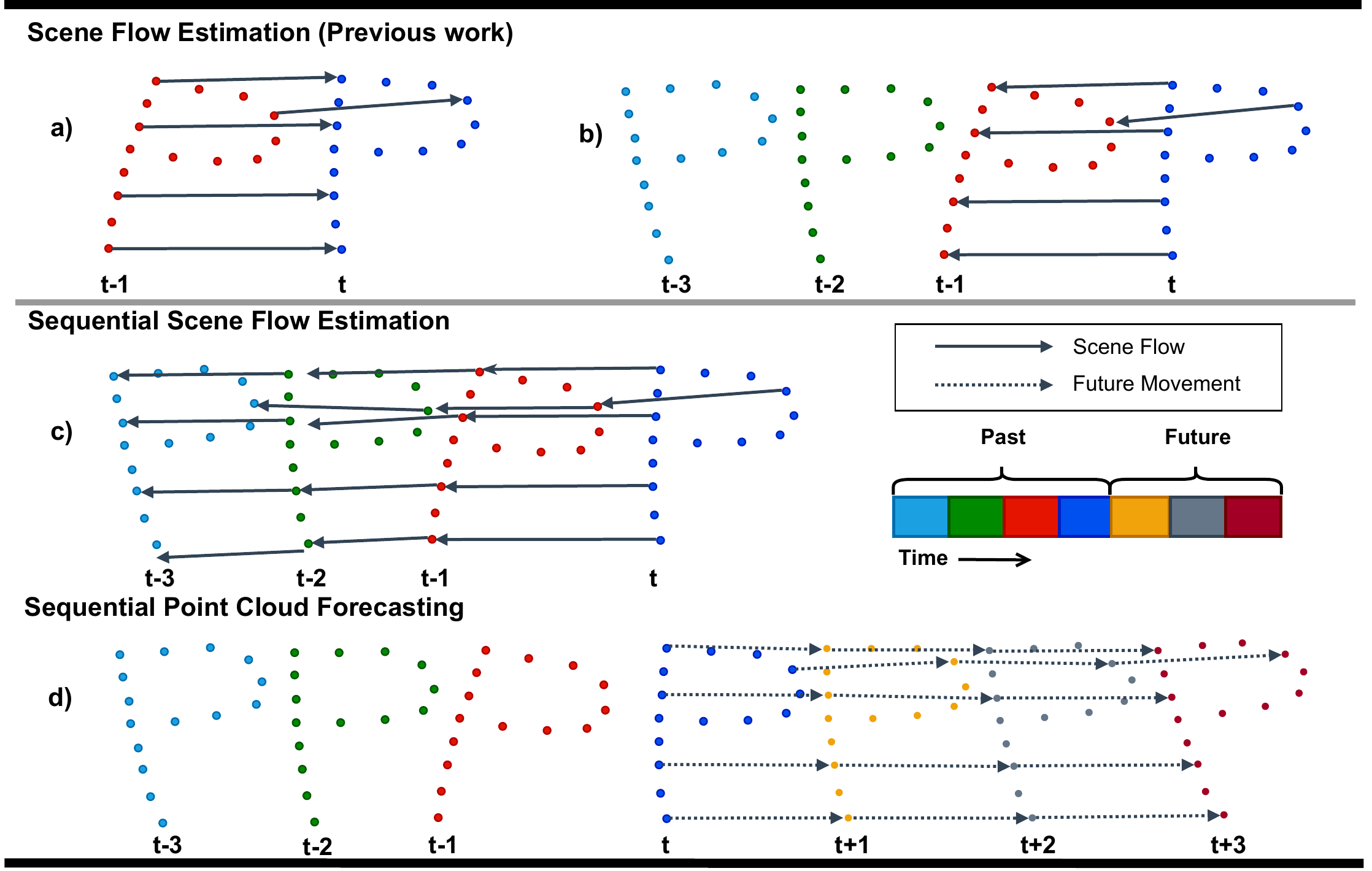}
    \caption{\textbf{Contrasting the proposed sequential scene flow estimation (SSFE) problem and standard scene flow estimation (SFE)}. a) SFE predicts relative motion between a single pair of frames and has been widely evaluated in prior work such as \citep{liu2019flownet3d,wu2019pointpwc,puy20flot}. b) The problem can be further elevated by utilizing preceding frames, as evidenced in MeteorNet \citep{liu2019meteornet}. c) \textbf{Our proposed SSFE problem} requires estimating 3D scene flow between \textit{multiple} adjacent frames. It requires processing an entire point cloud sequence, implying that multi-step spatiotemporal information is relevant for solving this task. It has been unexplored until now due to the lack of an appropriate benchmark with supervision for point cloud sequences and standardized training and evaluation protocols. Our proposed benchmark addresses this gap. d) We also include both supervised and self-supervised variants of the closely related task of sequential point cloud forecasting (SPF)~\citep{fan2019pointrnn,Weng2020_SPF2} in the new benchmark, which has likewise been difficult to study for the same reasons. This enables investigating whether, e.g., pre-training on SSFE aids SPF.
    }\label{fig:problem}
\end{figure*}
Autonomous agents need to understand 3D environments to ensure safe planning and navigation. A critical step is to perceive and predict the actions of entities such as vehicles, pedestrians, and cyclists. This requires learning rich embeddings of recorded data. Among various 3D geometric data representations, point clouds can accurately preserve the original geometric information
in 3D environments with less information loss compared to other representations such as voxels \citep{maturana2015voxnet}, or projected images \citep{wu2018squeezeseg}. This has led to the explosive growth in developing point cloud-based deep architectures, as evidenced in \citep{qi2017pointnet,qi2017pointnet++,su2018splatnet,landrieu2018large,liu2019flownet3d,dgcnn} and other tasks such as 3D semantic and instance segmentation \citep{wang2018sgpn,hou2019sis,yi2019gspn,pham2019jsis3d,zhao2020jsnet}, and 3D object detection \citep{qi2018frustum,gwak2020gsdn,shi2020pv,qi2020imvotenet,tang2020searching,gwak2020gsdn,yin2021center}. However, the community has paid less attention to the processing of dynamic point clouds in a spatiotemporal scene.
Unlike grid-based RGB images or videos, dynamic point clouds are unordered and irregular in the spatial dimension and can change drastically in the temporal dimension. 
The spatiotemporal processing of raw point cloud sequences remains an open challenge.

One fundamental 3D task is to understand the motion of a dynamically changing scene by estimating the scene flow between two consecutive point clouds (Fig. \ref{fig:problem}a). 
Despite receiving significant attention from the 3D community~\citep{liu2019flownet3d,gu2019hplflownet,wu2019pointpwc,puy20flot,mittal2020just}, most scene flow estimation (SFE) approaches have focused on inferring the relative motion of a given frame pair. Only one known study has considered using an input \textit{sequence} of point clouds~\citep{liu2019meteornet} (Fig. \ref{fig:problem}b), but they still only predict scene flow for a single pair of point clouds. Unlike previous work, we  instead consider a new sequence-to-sequence problem of obtaining a sequence of flow estimation or future movement conditioned on an input sequence of point clouds and conduct a through investigation with several contributions.

First, we introduce the sequential scene flow estimation (SSFE) task. 
Different from the standard SFE problem, models solving SSFE are evaluated on their ability to predict $T-1$ consecutive scene flows conditioned on an input sequence of $T$ point clouds. In SFE, models need only predict a single scene flow, whether the input is a pair of frames (Fig.~\ref{fig:problem}a) or a sequence~(Fig.~\ref{fig:problem}b).
SSFE is a non-trivial extension of SFE because current SFE methods are not equipped to extract multi-step spatiotemporal information from sequences and because SFE benchmarks do not support the training and evaluation of $T-1$ consecutive scene flows.

We also propose SSFE to help solve the challenging and related task of sequential point cloud forecasting (SPF, Fig. \ref{fig:problem}d)~\citep{Weng2020_SPF2}.
Unlike models for SSFE, models trained on SPF must predict a sequence of \emph{future} point clouds conditioned on a sequence of \emph{past} point clouds.
 SPF is still relatively unexplored. One study proposed a self-supervised architecture for this task but failed to adequately formalize the new task and evaluate the method~\citep{fan2019pointrnn}.  
Recently, another self-supervised architecture SPFNet~\citep{Weng2020_SPF2} was proposed as well as a formalization of self-supervised SPF.
They focus on trajectory forecasting and only provide a limited evaluation of the self-supervised SPF task.
In this work, we study the connection between our new SSFE problem and the supervised and self-supervised variants of SPF.

Second, we establish a method for solving SSFE called SPCM-Net (\textbf{S}equential \textbf{P}oint \textbf{C}loud \textbf{M}odeling \textbf{Net}work).
SPCM-Net extends a state-of-the-art coarse-to-fine SFE architecture~\citep{wu2019pointpwc} to exploit multi-step information, and differs from related models for sequential processing of point clouds by using a  set-to-set cost volume layer.
Specifically, SPCM-Net embeds the set-to-set cost volume layer within a recurrent cell at each scale of a feature pyramid.
This provides multi-scale spatiotemporal correlation information between neighboring point clouds which gets aggregated over time by an order-invariant recurrent unit.
SPCM-Net can be directly used to also solve SPF by appending a similarly-designed decoder to the architecture. Furthermore, we explore how pre-training on SSFE impacts performance when fine-tuning on SPF.

Our third contribution is to standardize training and evaluation protocols by introducing a rigorous benchmark consisting of several datasets for both the SSFE and SPF problems. No dataset, to the best of our knowledge, has been proposed to train and evaluate frame-wise scene flow estimation in point cloud sequences of lengths longer than two frames. To overcome this limitation, we take the popular synthetic FlyingThings3D dataset \citep{mayer2016large} and reconstruct point cloud sequences with multi-step ground truth scene flow to support the SSFE task. We repeat the same process on the newly released Virtual KITTI dataset \citep{gaidon2016virtual} for synthetic evaluation on traffic scenes. 
 We also process raw LIDAR sequences collected from the Argoverse dataset \citep{chang2019argoverse} for the self-supervised variant of the SPF task.
 We define suitable metrics  for  the  new  SSFE  problem  as  well  as  for SPF and adapt appropriate prior work \citep{qi2017pointnet,liu2019flownet3d,wu2019pointpwc,fan2019pointrnn,puy20flot} for comparison.

Experimental results on the new benchmark confirm the effectiveness of SPCM-Net on the new SSFE problem.
We demonstrate a clear advantage over pre-existing SFE methods due to the recurrent processing of point cloud sequences for learning scene dynamics. 
We also show that pre-training on SSFE followed by fine-tuning on the SPF task improves SPF performance significantly and establishes state-of-the-art performance compared to training from scratch.
Without additional pre-training, SPCM-Net achieves competitive performance on the SPF task compared to relevant prior work. 
In a control study on the popular KITTI SFE benchmark~\citep{menze2015object,liu2019meteornet}, we find that SPCM-Net's recurrent cost volume approach provides a stronger inductive bias for sequential point cloud processing than 4D convolution proposed by the state-of-the-art MeteorNet~\citep{liu2019meteornet}.

\subsection{Contributions}
We summarize the contributions of this paper as follows:

\begin{itemize}
    \item To the best of our knowledge, this is the first work to formally define the problem of sequential scene flow estimation (SSFE) for point cloud sequences.
     \item We propose the new SPCM-Net architecture for solving SSFE.
      SPCM-Net establishes the state-of-the-art performance on the new SSFE task by combining a set-to-set cost volume layer within a recurrent point cloud processing architecture. 
      \item We show that pre-training SPCM-Net on the proposed SSFE task improves the downstream performance on sequential point cloud forecasting.
    \item To aid future research we present a sequential point cloud benchmark consisting of two synthetic datasets and one real-world dataset. The benchmark standardizes metrics for both supervised and self-supervised task variants and provides multi-step ground truth motion annotations. 
\end{itemize}

In what follows, we describe SSFE and SPF problems (Section~\ref{sec:problem}) and then present our proposed method (Section~\ref{sec:method}).  Then, we introduce the proposed benchmark consisting of three new datasets along with appropriate evaluation metrics (Section~\ref{sec:dataset}). Experimental results are presented and discussed in Section \ref{sec:experiment}. Related work is discussed in Section \ref{sec:related}. We discuss findings, limitations, and future work in Section \ref{sec:discussion} and draw conclusions in Section \ref{sec:conclusion}.

\section{Problem Definitions}\label{sec:problem}

 \subsection{Sequential Scene Flow Estimation}  \label{sec:ssfe}
 
 SSFE requires capturing spatiotemporal interaction to estimate frame-wise motions of points in different frames (Fig. \ref{fig:problem}c). 
 Formally, the input is a sequence of $T$ consecutive point clouds $\boldsymbol{P}_{1:T}=\{ (\boldsymbol{C}_t, \boldsymbol{X}_t) \mid t=1,\dots, T \}$ with 3D point coordinates $\boldsymbol{C}_t \in \mathrm{R}^{N \times 3}$ and their corresponding features $\boldsymbol{X}_t \in \mathrm{R}^{N \times d}$, where $N$ and $d$ denote the number of points and feature dimensions, respectively.
 Given the sequence $\boldsymbol{P}_{1:T}$, the goal is to estimate the scene flow associated to each frame of the sequence (starting from the second frame).
 Denoting the predicted flows as $\widehat{\boldsymbol{S}}_{2:T}$ and the ground truth scene flows as $\boldsymbol{S}_{2:T}$ \footnote{In this paper, we define the ground truth scene flow as the motion from frame $T$ to frame $T-1$, a backward flow. It mostly follows the setting in \cite{liu2019meteornet} for a convenient comparison.},
 we want to find a function $f$ to compute $\widehat{\boldsymbol{S}}_{2:T} = f_{\text{ssfe}} (\boldsymbol{P}_{1:T})$ that minimizes the error defined as
 \begin{equation}
    E(f_{\text{ssfe}}, \boldsymbol{P}_{1:T}) = D_e (f_{\text{ssfe}} (\boldsymbol{P}_{1:T}), \boldsymbol{S}_{2:T}).
\end{equation} $D_e$ is generally instantiated as a mean square error between $\widehat{\boldsymbol{S}}_{2:T}$ and $\boldsymbol{S}_{2:T}$. The function $f_{\text{ssfe}}$ is modeled with a neural network suitable to point cloud sequences.

Prior work \citep{liu2019flownet3d,gu2019hplflownet,wu2019pointpwc,wang2020flownet3d++} has focused on the standard scene flow estimation problem between two consecutive frames. By contrast, SSFE requires processing sequences longer than two frames, which encourages the extraction of contextual information from all frames to achieve more accurate and robust estimation.

 \subsection{Sequential Point Cloud Forecasting} 
 
The SPF task is to process a given $T$-length sequence $\boldsymbol{P}_{1:T}$ and predict the most probable future point cloud sequence of length $K$, given by $\widehat{\boldsymbol{P}}_{T+1:T+K}=\{(\widehat{\boldsymbol{C}}_{T+k}, \widehat{\boldsymbol{X}}_{T +k}) \mid k=1,\dots,K\}$ (Fig. \ref{fig:problem}d):
\begin{equation}
    \widehat{\boldsymbol{P}}_{T+1:T+K} = \argmax_{\tilde{\boldsymbol{P}}_{T+1:T+K}} \Pr (\tilde{\boldsymbol{P}}_{T+1:T+K} \mid \boldsymbol{P}_{1:T}).
\end{equation}

 \noindent  The problem is highly non-trivial due to the complexity inherent in point cloud sequences, e.g., partial or full object occlusion, shape deformation, and scale variations.

 When ground truth point-wise motion is available, ground truth future frames can be generated from the input sequence $\boldsymbol{P}_{1:T}$ by adding 3D motion to the last point cloud of the sequence $\boldsymbol{P}_T$. 
 In this setting, the goal is to find a function $f_\text{spf}$ that minimizes the error between the predicted frames $\tilde{\boldsymbol{P}}_{T+1:T+K}=f_{\text{spf}} (\boldsymbol{P}_{1:T})$ and the ground truth future frames $\boldsymbol{P}_{T+1:T+K}$ defined as
 \begin{equation}
    E(f_{\text{spf}}, \boldsymbol{P}_{1:T}) = D_p (f_{\text{spf}} (\boldsymbol{P}_{1:T}), \boldsymbol{P}_{T+1:T+K}).
\end{equation} 
In this supervised setting, $D_p$ can be implemented as the mean square error. 

When ground truth point-wise motion is not available, the task can be approached in a self-supervised manner~\citep{fan2019pointrnn,Weng2020_SPF2}. Here,  pseudo-ground truth is generated by estimating nearest points to predicted points from the point cloud frame in the next timestep, and vice versa. Then, $D_p$ can be implemented as the Chamfer distance (CD), which is applied to every pair of predicted and future frames at each timestep (see Section \ref{sec:objectives} for a formal definition of CD).

The formulation of SPF above is a generalization of the same task considered in prior work~\citep{fan2019pointrnn,Weng2020_SPF2} as it admits both supervised and self-supervised approaches.
This eases the study of this problem within the context of our proposed benchmark which provides multi-step ground-truth motion annotations.

\section{Proposed Method}\label{sec:method}

In this section, we describe our proposed method for sequential scene flow estimation \emph{and} sequential point cloud forecasting.
Our model solves the defined tasks by exploiting several properties of point cloud sequences~\citep{liu2019meteornet,zhang2019cloudlstm}:

\begin{itemize}
     \item \textbf{Intra-frame order invariance}. Points within the same frame are arranged without a specific order. Any permutation applied to the points should not change the output of the model. 
     \item  \textbf{Inter-frame location variance}. Points at different timestamps may carry different spatial correlations. Such dynamic changes of spatial correlation should be captured by a model, i.e., changing the timestamp of a point should result in a different feature vector.
     \item \textbf{Spatiotemporal interaction between points}.\\ Points that are close spatially and temporally should be considered as neighboring points, from where local dependencies should be modeled. 
 \end{itemize}

\begin{figure*}
    \centering
 \includegraphics[width=\textwidth]{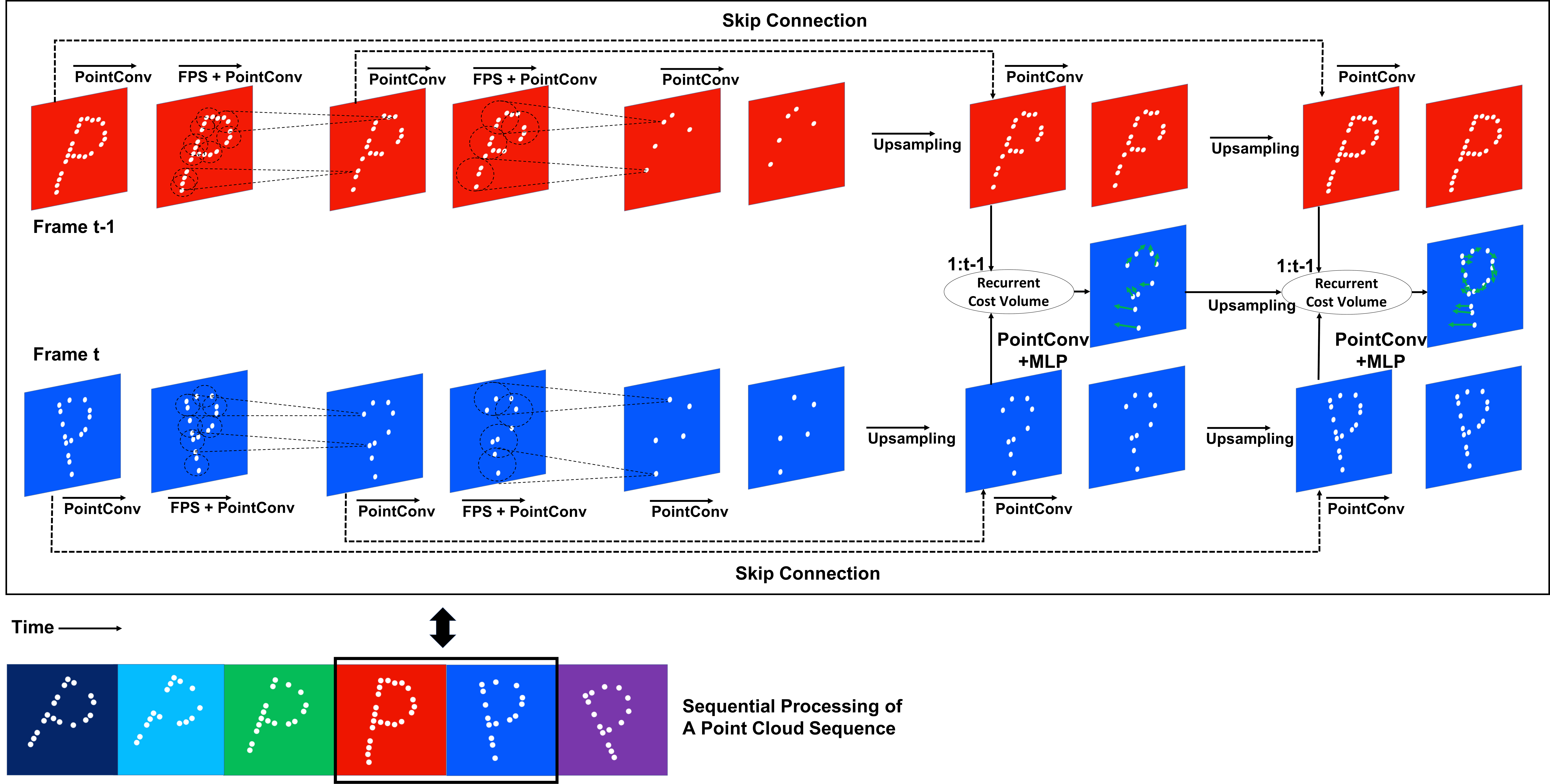}
    \caption{\textbf{The SPCM-Net architecture for sequential scene flow estimation}. We only show the operation of two frames here although SPCM-Net is designed to recurrently process a point cloud sequence. For each point cloud frame of the current frame fair,  we first encode it into a feature pyramid via a stack of PointConv \citep{wu2019pointconv}, multilayer perceptron (MLP), and Farthest Point Sampling (FPS) layers \citep{moenning2003fast,eldar1997farthest,qi2017pointnet++}. Then at each pyramid level $l$, its recurrent cost volume layer takes features of the current point cloud ($F_l$) as the input and updates its hidden states from $t-1$ to $t$. We concatenate the updated states, the upsampled coarse flow from pyramid level $l-1$, and feature $F_l$, followed by multiple PointConv~\citep{wu2019pointconv} and MLP layers, to generate a finer scene flow and the intermediate features. For simplicity, we omit one level and only visualize a three-level pyramid architecture. The future predictor is designed similarly and is described accordingly in Section \ref{sec:multiple_prediction}.}
    \label{fig:pcmnet}
\end{figure*}
\subsection{Network Design}
Due to the nature of SSFE and SPF, models must have a capability of \textit{accurately capturing multi-step spatiotemporal information from point cloud sequences}. Existing SFE approaches will not adapt to these tasks because they are originally designed to handle frame pairs.
Without a temporal receptive field spanning the input point cloud sequence, they are likely to fail to exploit multi-step information. Inspired by this, we propose SPCM-Net to recurrently process each pair of frames in a sequence and aggregate features in a spatiotemporal fashion.  

The architecture of SPCM-Net for the SSFE task is visualized in Fig. \ref{fig:pcmnet}. It consists of multiple modules including an intra-frame feature pyramid (IFFP) module, an inter-frame spatiotemporal correlation (IFSC) module, and a multi-scale coarse-to-fine  prediction (MCP) module. The IFFP module encodes each point cloud frame into a feature pyramid that captures local spatial information at different scales. The IFSC module recurrently processes each pair of frames in the sequence and fuses features with past information, which we will describe in Section \ref{sec:ifsc}. The MCP module generates multi-scale prediction at each timestep  based on features from IFFP and IFSC modules. Depending on the specific task, the MCP module is to either estimate scene flows or predict future point movements.

We now describe each module of the SPCM-Net architecture for the SSFE task and discuss key ingredients to our approach. The architecture for the SPF task is similarly designed and described in Section \ref{sec:multiple_prediction}.

\subsubsection{Intra-frame Feature Pyramid} \label{sec:intra}
We can not directly apply conventional convolution operators to point clouds because they are irregular and orderless. Therefore, we follow PointPWC-Net architecture \citep{wu2019pointpwc} to utilize the PointConv layer \citep{wu2019pointconv} to capture local spatial information within each point cloud. This generates pyramidal features by hierarchically sampling each point cloud via multiple Farthest Point Sampling (FPS) \citep{moenning2003fast,eldar1997farthest,qi2017pointnet++} layers. Each pyramid captures the local geometric structure within its receptive field. The pyramid features are further aggregated into larger units to generate higher-level features. We repeat the process until reaching a demanding number of pyramid levels, e.g., five~\citep{wu2019pointpwc}.

\subsubsection{Inter-frame Spatiotemporal Correlation}\label{sec:ifsc}

To model spatiotemporal correlation between point cloud frames, we would like to exploit the local dependencies of neighboring frames to capture a larger temporal receptive field across the whole sequence length. A straightforward solution would be directly applying 4D convolutions on the voxelized sequence following \citep{choy20194d}. However, this requires extensive computation. Furthermore, quantization errors during voxelization may cause performance drops when tackling problems requiring precise measurement, e.g., point-wise scene flow estimation. A promising solution would be instead borrowing designs from sequence modeling to construct a recurrent model customized for point cloud sequences.

PointPWC-Net uses a cost volume for computing the scene flow between two consecutive frames. In our model, within each pyramid level, the pyramid features generated from two frames are combined to compute a cost volume to obtain spatiotemporal correlation. Although we could simply repeat this computation between every two adjacent frames to predict scene flow for a given point cloud sequence, this ignores multi-step information in preceding frames which could lead to more accurate estimation. Instead, we combine the cost volume  for pairwise frame modeling with a recurrent neural unit similar to long short-term memory (LSTM)~\citep{hochreiter1997long,graves2012supervised}. The spatiotemporal features produced by the cost volume are fused with past information, producing a smoother estimation of 3D scene flow for each point. 
This is important for both the SSFE and SPF tasks. 
The detailed description of the \emph{recurrent} cost volume (RCV) layer can be found in Section \ref{sec:architectural}.

\subsubsection{Multi-scale Coarse-to-fine Prediction}\label{sec:multiple_prediction}

Inspired by scene flow approaches \citep{revaud2015epicflow,sun2018pwc,gu2019hplflownet,wu2019pointpwc}, we adopt a \textit{coarse-to-fine} prediction approach where the current scene flow prediction is initialized with estimated flows from a preceding prediction. We establish this by plugging RCV layers into the feature pyramid defined in Section~\ref{sec:intra}. At each pyramid level, the RCV layer builds a spatiotemporal correlation between downsampled versions of the current pair of point cloud frames in the sequence via FPS \citep{eldar1997farthest,moenning2003fast,qi2017pointnet,yu2018pu,li2018pointcnn,qi2019deep}. At the top level is the original input point clouds. The bottom level contains the fewest points, which generates the coarsest scene flows. We upsample these flows with respect to points in the higher level via the widely used inverse distance weighted interpolation \citep{qi2017pointnet++} and make a further refinement.

\textbf{Sequential flow estimator}. At each pyramid level $l$ of the current timestep $t$, the RCV takes features of the current point cloud ($F_l^t$) as the input and updates its hidden states from $t-1$ to $t$. We concatenate the updated states, the upsampled coarse flow from the pyramid level $l-1$, and features $F_l$, followed by a stack of PointConv~\citep{wu2019pointconv} and multilayer perceptron (MLP) layers, to generate a finer scene flow and the intermediate features. 

Formally, let $SF_{l,t}$ be the estimated flow at level $l$ of timestep $t$, $\boldsymbol{P}_{t,l}$ be the point cloud at level $l$ of timestep $t$, we upsample the estimated coarse flow $SF_{l-1,t}$ at level $l-1$ with respect to  $\boldsymbol{P}_{t,l}$ to obtain the upsampled coarse flow. For each point $p_i^{l,t}$ in the fine level point cloud $\boldsymbol{P}_{t,l}$, we find its K nearest neighbors $N(p_i^{l,t})$ in the coarse level point cloud $\boldsymbol{P}_{t,l-1}$. Each scene flow $\hat{sf}_{l,t}^i$ in $\hat{SF}_{l,t}$ for finer level $l$ is computed via inverse distance weighted interpolation:
\begin{equation}
    \hat{sf}_{l,t}^i = \frac{\sum_{j=1}^K \omega (p_i^{l,t}, p_j^{l-1,t}) SF^{l-1,t}_j }{\sum_{j=1}^K \omega (p_i^{l,t}, p_j^{l-1,t})}
\end{equation} 
where $\omega (p_i^{l,t}, p_j^{l-1,t}) = \frac{1}{d(p_i^{l,t}, p_j^{l-1,t})}$ and $p_j^{l-1,t} \in N(p_i^{l,t})$. We used the Euclidean distance as the distance metric $d(p_i^{l,t}, p_j^{l-1,t})$. The estimated flow $SF_{l,t}$ is further obtained by concatenating $\hat{SF}_{l,t}$, $F_l^t$ and the updated state of RCV ($\boldsymbol{H}_t$) and feeding them to a stack of PointConv~\citep{wu2019pointconv} and MLP layers:
\begin{equation}
    SF_{l,t} = MLP(PointConv(\widehat{SF}_{l,t};F_l^t;\boldsymbol{H}_t).
\end{equation}

We repeat the process for all pyramid levels. The final estimated scene flow at time $t$ is $SF_{L,t}$. By doing so, we have modeled a point cloud sequence via multiple RCV layers at different pyramid levels and across different timesteps. This design allows exploiting stronger spatiotemporal correlation in sequences, which we verify in the experiment section.

\textbf{Sequential future predictor}. 
The architecture from the SSFE task can be adapted to support the SPF task by treating it as an encoder and adding a decoder. The encoder digests input point cloud frames while the decoder predicts the future movement of the last input point cloud $\boldsymbol{P}_T=\ (\boldsymbol{C}_T, \boldsymbol{X}_T)$. Specifically, the encoder consumes the input point cloud sequence frame-by-frame and keeps updating the states of RCV layers till $\boldsymbol{P}_T$. The obtained states will initialize the states for the decoder. To simplify the problem, we predict the displacements $\Delta \boldsymbol{P}$ between points of the current timestep and the next timestep rather than directly reconstructing future point coordinates from scratch, which is generally more difficult. We feed the predicted point cloud frame into the model to interact with the states of the RCV layers and generate the next point cloud. We repeat this operation until the prediction step reaches $K$.

\subsection{Recurrent Cost Volume Layer} \label{sec:architectural}

This section describes the recurrent cost volume in detail. We begin by providing a preliminary introduction to the cost volume to build the necessary background. 
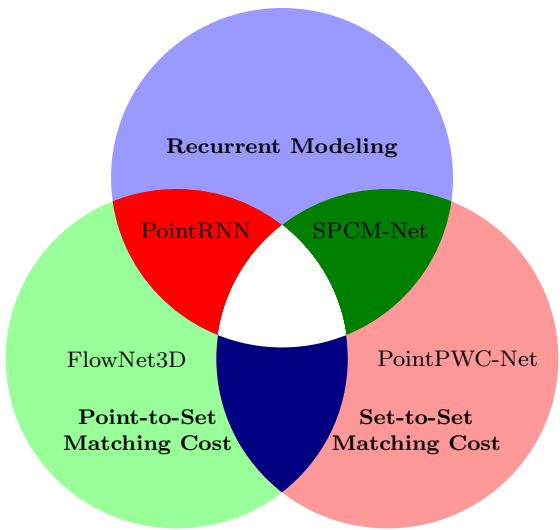
\begin{figure}
    \centering
    \begin{tikzpicture}

    \node [venn circle=green, minimum size=4.5cm, label={[label distance=-2.5cm]180:FlowNet3D}] (A) at (-150:1.6cm) {};
    \node [venn circle=blue, minimum size=4.5cm, label={[label distance=-2cm]90:}] (B) at (90:1.6cm) {};
    \node [venn circle=red,  minimum size=4.5cm, label={[label distance=-2.5cm]0:PointPWC-Net}] (C) at (-30:1.6cm) {};
\fill[red,
          intersection segments={
            of=green and blue,
            sequence={R2--L2}
          }]; 
\fill[blue!50!black,
          intersection segments={
            of=red and green,
            sequence={R1--L2--R0}
          }];
\fill[green!50!black,
          intersection segments={
            of=red and blue,
            sequence={R2--L2}
          }];
\path [
    name path=rag,
    intersection segments={
        of=green and red,
    }];
\fill[white,intersection segments={of=blue and rag,sequence=R2--B1}]  
    [intersection segments={of=rag and blue, sequence={--R2}}];               
    \node[font=\footnotesize,left, xshift=4mm, yshift=5mm] at (barycentric cs:A=1/2,B=1/2 ) {PointRNN}; 
    \node[font=\footnotesize,right, xshift=-4mm, yshift=5mm] at (barycentric cs:B=1/2,C=1/2 ) {SPCM-Net}; 
\node[font=\footnotesize] at (90:2cm) {\textbf{Recurrent Modeling}};
\node[font=\footnotesize] at (225:2.5cm){\begin{tabular}{c} \textbf{Point-to-Set} \\ \textbf{Matching Cost} \end{tabular} };
\node[font=\footnotesize] at (315:2.5cm){\begin{tabular}{c} \textbf{Set-to-Set} \\ \textbf{Matching Cost} \end{tabular} };
\end{tikzpicture}  
    \caption{A comparison between different model architectures. The key operation in PointPWC-Net \citep{wu2019pointpwc} is the \textit{cost volume}, which we consider as \textit{the set-to-set matching cost}, in contrast to \textit{the point-to-set matching cost} (the \textit{flow embedding} layer) proposed by FlowNet3D \citep{liu2019flownet3d}. PointRNN \citep{fan2019pointrnn} incorporates the \textit{flow embedding} layer of FlowNet3D \citep{liu2019flownet3d} into a recurrent unit for predicting future point clouds while the proposed SPCM-Net extends PointPWC-Net \cite{wu2019pointpwc} by introducing the \textit{recurrent cost volume} that combines the cost volume for pairwise frame modeling with a recurrent neural unit similar to long short-term memory (LSTM)~\citep{hochreiter1997long,graves2012supervised}.}
    \label{fig:comparison}
\end{figure}

We first introduce the learnable matching cost between two consecutive point clouds following  PointPWC-Net~\citep{wu2019pointpwc}. Formally, given two points $\boldsymbol{p}_{t-1}^i=\ (\boldsymbol{c}_{t-1}^i, \boldsymbol{x}_{t-1}^i) \in \boldsymbol{P}_{t-1} $ and $\boldsymbol{p}_t^j=\ (\boldsymbol{c}_t^j, \boldsymbol{x}_t^j) \in \boldsymbol{P}_t$ with 3D point coordinates and  their  corresponding  features, the matching cost between $\boldsymbol{p}_t^j$ and $\boldsymbol{p}_{t-1}^i$ is defined as
\begin{equation}
    \label{eq:cost}
        \text{Cost}(\boldsymbol{p}_t^j, \boldsymbol{p}_{t-1}^i) = \phi_{\text{MLP}} (\boldsymbol{c}_{t-1}^i-\boldsymbol{c}_t^j, \boldsymbol{x}_{t-1}^i, \boldsymbol{x}_t^j).  
\end{equation} Here, the feature vectors of two points and the directional difference between their spatiotemporal positions are passed to an MLP. Note that $\boldsymbol{p}_{t-1}^i$ comes from a neighboring
point set of $\boldsymbol{p}_t^j$ based on spatiotemporal distance or feature similarity.

However, it has been shown that this pure point-to-point matching cost is sensitive to outliers \citep{wu2019pointpwc}. The \textit{flow embedding} layer proposed by \citep{liu2019flownet3d} partly addresses it by aggregating flow votes from neighboring points. Specifically, for a given point $\boldsymbol{p}_{t}^j$, they finds its neighboring points at timestep $t-1$ via ball query. These points are considered as multiple soft correspondence points for $\boldsymbol{p}_{t}^j$ and are utilized to obtain multiple matching costs defined in Equation \ref{eq:cost}. Matching costs are further aggregated via the max-pooling. However, motion information can be lost due to the max-pooling operation. To obtain a more robust and stable matching cost, a preferable approach is to aggregate matching costs in a manner similar to the patch-to-patch approach in optical flow~\citep{hosni2012fast,sun2018pwc}. 
This motivates us to choose the \textit{cost volume} in PointPWC-Net \citep{wu2019pointpwc} to describe point motion. We consider the cost volume as \textit{the set-to-set matching cost}, in contrast to \textit{the point-to-set matching cost} (the \textit{flow embedding} layer).
The leap from point-to-set to set-to-set matching costs is exactly prefigured in the move from the softmax to the softassign cost in earlier point matching \citep{chui2003new}. We provide a comparison between our proposed SPCM-Net and several model architectures including PointRNN \citep{fan2019pointrnn}, FlowNet3D \citep{liu2019flownet3d}, and PointPWC-Net \citep{wu2019pointpwc} in Fig. \ref{fig:comparison}.

Formally, the cost volume for $\boldsymbol{p}_t^j$ is defined as 
\begin{equation}
\begin{split}
    \text{CV} (\boldsymbol{p}_t^j)  & = \sum_{\boldsymbol{p}_t^k \in M (\boldsymbol{p}_t^j)} \omega_M (\boldsymbol{p}_t^k, \boldsymbol{p}_t^j) \times \\ 
    & \sum_{\boldsymbol{p}_{t-1}^i \in N (\boldsymbol{p}_t^k)}  \omega_N ( \boldsymbol{p}_{t-1}^i, \boldsymbol{p}_t^k) \text{ Cost} (\boldsymbol{p}_{t-1}^i, \boldsymbol{p}_t^k) \label{eq:cost_volume}
\end{split}
\end{equation}
\vspace{-5mm}
\begin{align}    
    &\omega_M (\boldsymbol{p}_t^k, \boldsymbol{p}_t^j) = \text{MLP} (\boldsymbol{c}_t^k-\boldsymbol{c}_t^j) \\
    & \omega_N ( \boldsymbol{p}_{t-1}^i, \boldsymbol{p}_t^k) = \text{MLP} (\boldsymbol{c}_{t-1}^i-\boldsymbol{c}_t^k)
\end{align}
This requires finding a spatial neighboring point set $M (\boldsymbol{p}_t^j)$ around $\boldsymbol{p}_t^j$ in $\boldsymbol{P}_t$.
Then for each point $\boldsymbol{p}_t^k \in M (\boldsymbol{p}_t^j)$, we find a spatiotemporal neighboring point set $N (\boldsymbol{p}_t^k)$ around $\boldsymbol{p}_t^k$ in $\boldsymbol{P}_{t-1}$ (across time). The interaction between these points is modeled by two directional vectors obtained via convolutional operations $ \omega_M (\boldsymbol{p}_t^k, \boldsymbol{p}_t^j)$ and $\omega_N ( \boldsymbol{p}_{t-1}^i, \boldsymbol{p}_t^k)$, and their matching costs. Both the spatial neighboring point set  $M (\boldsymbol{p}_t^j)$ and  $N (\boldsymbol{p}_t^k)$ can be obtained by conducting an efficient GPU-based ball query that finds all points within a radius to the query point or the K-nearest neighbor search that finds a fixed number of points that are the closest.

\begin{figure}
    \centering
    \includegraphics[width=0.5\textwidth]{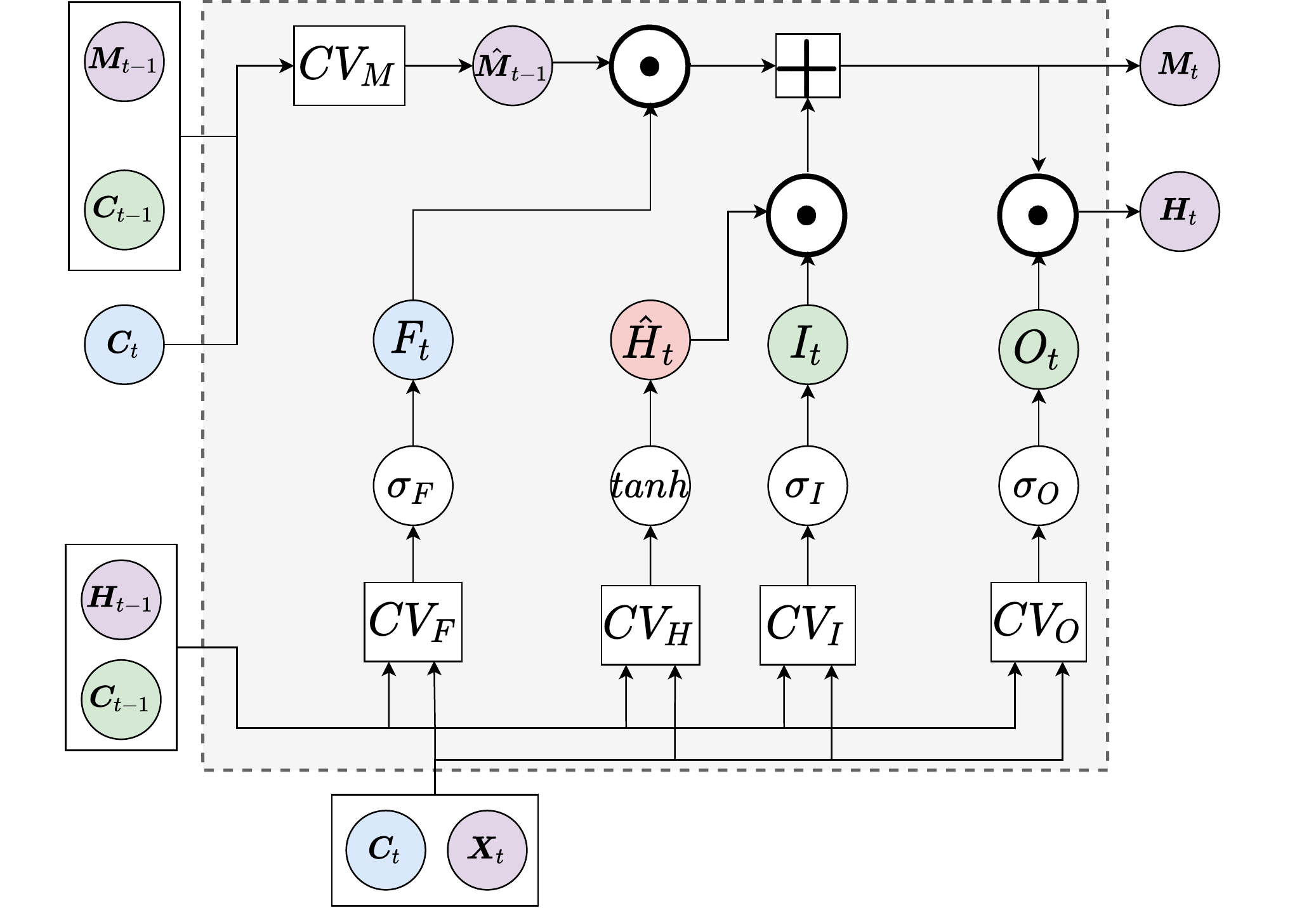}
    \caption{\textbf{Recurrent Cost-Volume Layer}. At each timestep $t$, it takes the current point locations $\boldsymbol{C}_t$ and the associated features $\boldsymbol{X}_t$ as the inputs. They will interact with recurrent cost volume memory states $\boldsymbol{C}_{t-1}$, $\boldsymbol{H}_{t-1}$, and $\boldsymbol{M}_{t-1}$ via multiple gates.}
    \label{fig:rcv}
\end{figure}

Now we show how to embed the cost volume into a recurrent unit for point cloud sequences (Fig. \ref{fig:rcv}).

\textbf{Inputs:} To maintain spatial structure, our hidden states maintain both the point coordinates and the associated features. At timestep $t$, both $\boldsymbol{C}_t$ and $\boldsymbol{X}_t$ will be fed into the RCV layer as the input.

\textbf{Initialization:} Accordingly, the states of the RCV layer are extended to $\boldsymbol{C}_{t-1}$, $\boldsymbol{H}_{t-1}$ and $\boldsymbol{M}_{t-1}$ to track the most recent historic point locations and memory states. For notation clarity, since we already are using $\boldsymbol{C}_t$ to denote the coordinates of points in the point cloud, we will use $\boldsymbol{M}_{t}$ to refer to the recurrent cell state. The hidden and cell states are zero-initialized at time $t=0$.

\textbf{Order-Invariance:} 
Since the point sets can be ordered completely differently across time and objects may appear or disappear from view, there is no guarantee that a one-to-one mapping between neighboring frames exists.
Therefore, we use the cross-frame neighborhood query within the cost volume operation to perform updates to the hidden and cell states ($\boldsymbol{H}_{t-1}$ and $\boldsymbol{M}_{t-1}$), making them invariant to any changes to the order of points or to the addition of new points. 

\textbf{Update:} 
Let
\begin{align}
\text{CV} (\boldsymbol{P}_t ; \boldsymbol{P}_{t-1}) =\text{CV} (\boldsymbol{C}_t, \boldsymbol{X}_t; \boldsymbol{C}_{t-1}, \{\boldsymbol{H}_{t-1}, \boldsymbol{M}_{t-1}\})
\end{align}
be the cost volume for all points in timestep $t$ with respect to the point cloud at timestep $t-1$. 
The core component of the update operator is a recurrent unit similar to the LSTM cell, where we replace the fully connected layers with the cost volume. 
The relevant update equations are:

\begin{align}
    I_t &= \sigma_I (CV_I(\boldsymbol{C}_t, \boldsymbol{X}_t; \boldsymbol{C}_{t-1}, \boldsymbol{H}_{t-1}), \\
     F_t &= \sigma_F (CV_F(\boldsymbol{C}_t, \boldsymbol{X}_t; \boldsymbol{C}_{t-1}, \boldsymbol{H}_{t-1}), \\
    O_t &= \sigma_O (CV_O(\boldsymbol{C}_t, \boldsymbol{X}_t; \boldsymbol{C}_{t-1}, \boldsymbol{H}_{t-1}), \\
    \hat{\boldsymbol{M}}_{t-1} &=  CV_M(\boldsymbol{C}_t, \text{None}; \boldsymbol{   C}_{t-1}, \boldsymbol{M}_{t-1}), \\
    \hat{\boldsymbol{H}}_{t} &= \text{tanh}( CV_H(\boldsymbol{C}_t, \boldsymbol{X}_t; \boldsymbol{C}_{t-1}, \boldsymbol{H}_{t-1})), \\
     \boldsymbol{M}_t &= F_t\odot \hat{\boldsymbol{M}}_{t-1} +  I_t \odot \hat{\boldsymbol{H}}_{t}, \\ 
     \boldsymbol{H}_t &= O_t\odot  \boldsymbol{M}_t,  
\end{align} where the operator $\odot$  denotes the Hadamard product. $CV_I$, $CV_F$, and $CV_O$ denote the cost volume operations for the $input$, $forget$ and $output$ gates, respectively. 
The ``None'' in the cost volume operation applied to the cell state $\boldsymbol{M}_{t-1}$ represents that we do not use any input features when performing the neighborhood query on the cell state ($x_t^j$ in Eq.~\ref{eq:cost} is ignored).
Then  $\boldsymbol{\hat{M}}_{t-1}$  is modulated via the forget gate $F_t$ and is further aggregated with the new memory $\hat{\boldsymbol{H}}_{t}$ passed to the input gate $I_t$ to obtain the latest memory state $\boldsymbol{M}_{t}$. The latest hidden state $\hat{\boldsymbol{H}}_{t}$ is obtained by conditionally deciding what to output from $\boldsymbol{M}_{t}$ controlled by the output gate $O_t$. Note that $\boldsymbol{C}_t$ and $\boldsymbol{H}_{t}$ can be used together as the input features of downstream tasks.

\subsection{Learning Objectives} \label{sec:objectives}

\textbf{SSFE}.
 When ground truth scene flows are available, we adapt the multi-scale loss function used in PWC-Net \citep{sun2018pwc} and PointPWC-Net \citep{wu2019pointpwc} and extend it to handle point cloud sequences. Given the predicted  scene flow $\boldsymbol{SF}_{t,l}$  at the pyramid level $l$ from timestep $t$ and its ground truth scene flow $\boldsymbol{GF}_{t,l}$. The objective function is specified as
\begin{equation}
    L_{\text{supervised}}^{\text{SSFE}} = \sum_{t=2}^T \sum_{l=1}^L \alpha_l  || \boldsymbol{SF}_{t,l} - \boldsymbol{GF}_{t,l} ||_2^2. \label{eq:learning}
\end{equation} Occluded points are not considered by masking them out from gradient computation and weight updating. As done previously~\citep{wu2019pointpwc}, we use a set of hyper-parameters $\{\alpha_l \mid l=1,\dots, L \}$ to balance the importance of losses from different pyramid levels.

In this study, we do not explore self-supervised SSFE as this would require non-trivial innovation to develop a suitable training objective. 
Recent work~\citep{mittal2020just} has shown progress on self-supervised SFE which suggests that an extension to SSFE is possible.

\noindent \textbf{SPF.}
When ground truth point-wise motion is available, the learning objective is similar to Equation \ref{eq:learning}. We compute the difference between the predicted future frames and the  future ground truth frames derived from ground truth scene flow. Denote $\boldsymbol{P}_{t,l}$ and $\widehat{\boldsymbol{P}}_{t,l}$ as ground truth and predicted frames at the pyramid level $l$ from timestep $t$. The objective function is specified as
\begin{equation}
    L_{\text{supervised}}^{\text{SPF}} = \sum_{t=T+1}^{T+K} \sum_{l=1}^L \alpha_l  || \widehat{\boldsymbol{P}}_{t,l}- \boldsymbol{P}_{t,l} ||_2^2.
\end{equation} 
In reality, it is often difficult and expensive to obtain ground truth scene flows for real-world point cloud sequences and therefore few scene flow datasets are available. To avoid relying on the availability of ground truth scene flows, we can also define a self-supervised learning objective to train the model. We adopt the Chamfer Distance (CD) to compute the difference between predicted sequences and actual future sequences~\citep{Weng2020_SPF2}, which allows us to approximate the ground truth scene flow and guide the model learning. The CD is defined as
\begin{equation}\label{eq:chamfer_distance}
\scriptsize
D_{CD}(\boldsymbol{P}_t, \widehat{\boldsymbol{P}}_t) = \sum_{\boldsymbol{p} \in \boldsymbol{P}_t} \min_{\hat{\boldsymbol{p}} \in \widehat{\boldsymbol{P}}_t} \|\boldsymbol{p}-\hat{\boldsymbol{p}}\|^2 +  \sum_{\hat{\boldsymbol{p}} \in \widehat{\boldsymbol{P}}_t} \min_{\boldsymbol{p} \in \boldsymbol{P}_t} \|\hat{\boldsymbol{p}}-\boldsymbol{p}\|^2, 
\end{equation} where $\boldsymbol{P}_t$ and $\widehat{\boldsymbol{P}}_t$ are ground truth and predicted frames. We apply Equation \ref{eq:chamfer_distance} to all the future frames:
\begin{equation}
    L_{\text{self-supervised}}^{\text{SPF}} = \sum_{t=T+1}^{T+K} \sum_{l=1}^L \alpha_l  \mathcal{D}_{CD}( \widehat{\boldsymbol{P}}_{t,l}, \boldsymbol{P}_{t,l}).
\end{equation}
We do not explore advanced techniques widely used in the scene flow community for self-supervised SPF, e.g., Laplacian regularization or local smoothness \citep{wu2019pointpwc,pontes2020scene}, to further improve the performance. It guarantees a relatively fair comparison to baseline methods with simple learning objectives. 

\section{Datasets and Metrics}\label{sec:dataset}

In this section, we will describe datasets and the evaluation metrics designed for new tasks.

\textbf{Limitations of existing benchmarks}. Most existing benchmarks \citep{mayer2016large,menze2015object} focus on SFE between two consecutive point cloud frames with ground truth annotations, which are widely adopted in recent state-of-the-art approaches \citep{liu2019flownet3d,gu2019hplflownet,wu2019pointpwc}. An extension of the KITTI scene flow dataset to short sequences for flow estimation of the last input frame has been considered \citep{liu2019meteornet}. However, this does not meet the requirement of multi-step scene flow annotations necessary for supervised SSFE and SPF.

\textbf{New benchmarks.} Therefore, to evaluate SSFE and SPF, new datasets are needed to help systematically analyze novel methods.
For supervised SSFE and SPF we adapt two synthetic yet challenging datasets: FlyingThings3D \citep{mayer2016large} and Virtual KITTI \citep{gaidon2016virtual}. We generate ground truth annotations for point cloud sequences that could be useful to both SSFE and SPF tasks.
For self-supervised SPF we extract sequences from the real-world Argoverse dataset \citep{chang2019argoverse}.

\begin{figure}[hbt!]
    \centering
    \includegraphics[height=7.5cm,width=0.48\textwidth]{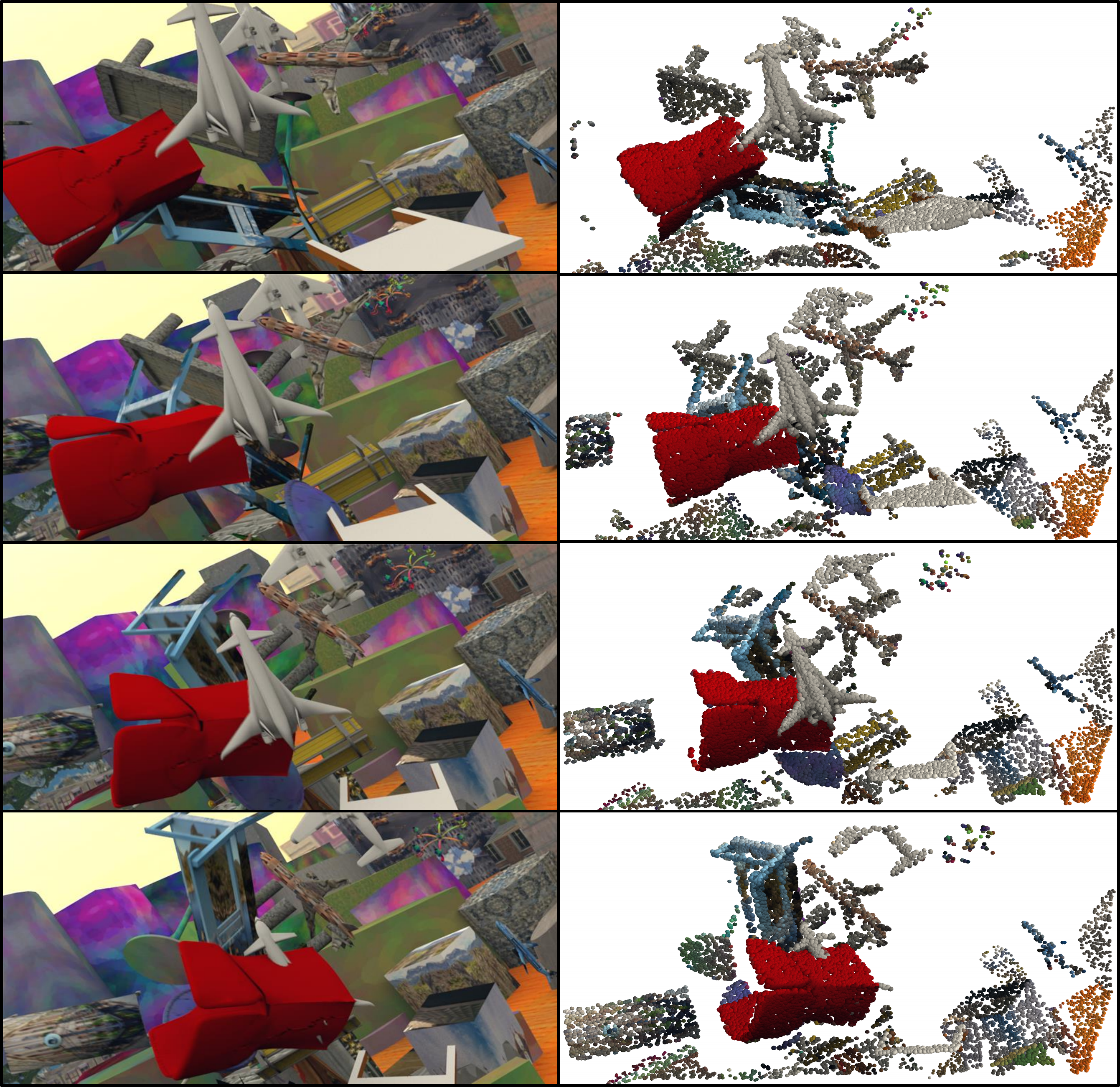}
    \caption{\textbf{One example sequence from our created SFT3D dataset} (the frame number increases from top to bottom). \textbf{Left}. The original frames of FlyingThings3D dataset \citep{mayer2016large}. \textbf{Right}. The reconstructed point cloud sequences. Notice that existing scene flow approaches only take a pair of frames as the input while in this paper we focus on handling point cloud sequences. Best viewed in color.}
    \label{fig:sft3d}
\end{figure}

\subsection{Sequential FlyingThings3D (SFT3D) Dataset}
FlyingThings3D \citep{mayer2016large} is the first large-scale synthetic dataset proposed for training deep learning models on scene flow estimation.
It contains videos all with a frame length of 10, rendered from scenes by randomly moving objects from the ShapeNet dataset \citep{chang2015shapenet}.  However, it does not provide point cloud sequences directly. Therefore, we reconstructed point clouds and 3D scene flows based on the ground-truth disparity maps, maps of disparity change, optical flows, and the provided camera parameters. We follow \citep{liu2019flownet3d,gu2019hplflownet} and only maintain points with a depth of less than $35$ meters.

Our SFT3D dataset can be used for evaluating both SSFE and SPF. In detail:
\begin{itemize}
    \item \textbf{SSFE}. We use the first six frames as the input for SSFE. Models must predict all the scene flows starting from frame 2 to frame 6. All these frames are provided with ground truth scene flows. For input frames, we randomly sample a fixed number of points (e.g., $2,048$ points) for each frame in a non-corresponding manner, meaning a point of a certain frame may not necessarily find its corresponding point in the subsequent frame. 
    \item \textbf{SPF}. We take the first six frames as the input while using the rest of the frames as ground truth. For all points in future frames, we find and track the future movement of points sampled in the last input frames (the $6^{\text{th}}$ input frames) along the whole prediction period to obtain the ground truth motions. 
\end{itemize}

The proposed SFT3D dataset is challenging, e.g., it contains points of occluded scene flows. Similar to the preparation of \citep{liu2019flownet3d}, occluded points are present in both the input and output of a designed model. However, they are not considered during performance evaluation or included in training losses. We remove sequences where all points are completely occluded in any frame. A visual example can be found in Fig. \ref{fig:sft3d}.

\begin{table*}[hbt!]
\centering
\caption{A summary of the created datasets with X forward, Y left,
and Z up}\label{tab:summary}
\resizebox{\textwidth}{!}{
\begin{tabular}{l|c|c|c|c|c|c|c|c}
\hline\noalign{\smallskip}
\multirow{2}{*}{Dataset} & \multirow{2}{*}{Frame Sampling}                   & \multirow{2}{*}{X-Y-Z Point Range}             & \multirow{2}{*}{Points Per Frame} & \multirow{2}{*}{\#Train} & \multirow{2}{*}{\#Validation} & \multirow{2}{*}{\#Test} & \multirow{2}{*}{SSFE} & \multirow{2}{*}{SPF} \\ 
                              &                                                   &                                                &                                   &                         &                              &                        &             &      \\ \noalign{\smallskip}\hline\hline
STF3D                         &  n/a         & $X\leq35$                                        & 2,048                             & 4,020                   & 446                          & 871                    & \checkmark      &    \checkmark      \\ \hline
VKS                           & Sample  10 frames per second & $X\leq35$                                        & 2,048                             & 2,175                   & N/A                          & 1,175                  &         \checkmark         &     \checkmark     \\ \hline
SAG                           & Sample 10 frames per second  & $[-32, 32] \times [-8, 8] \times [-\infty, 2]$ & 2,048                             & On-the-fly              & N/A                          & 1,200                  &     $\text{\sffamily X}$           &     \checkmark     \\ \hline
\end{tabular}
}
\end{table*}

\subsection{Virtual KITTI Sequence (VKS) Dataset} \label{sec:vks_dataset}
The Virtual KITTI dataset uses a game engine to recreate real-world videos from the KITTI tracking benchmark \citep{Geiger2012CVPR}. Due to recent improvement in lighting and post-processing of the Unity game engine, an improved dataset called the Virtual KITTI 2 dataset is released to be more photo-realistic and better-featured. It provides images from a stereo camera with new supports for forward and backward optical flow, forward and backward scene flow, and available camera parameters. We consider vehicles as objects of interest as they are the main dynamic objects in a traffic scene, i.e., trucks, cars, and vans. We obtain 3D point locations by projecting 2D pixel positions (in meters) to the 3D space based on the camera parameters. 
\begin{figure}
\centering
\includegraphics[height=8cm,width=0.48\textwidth]{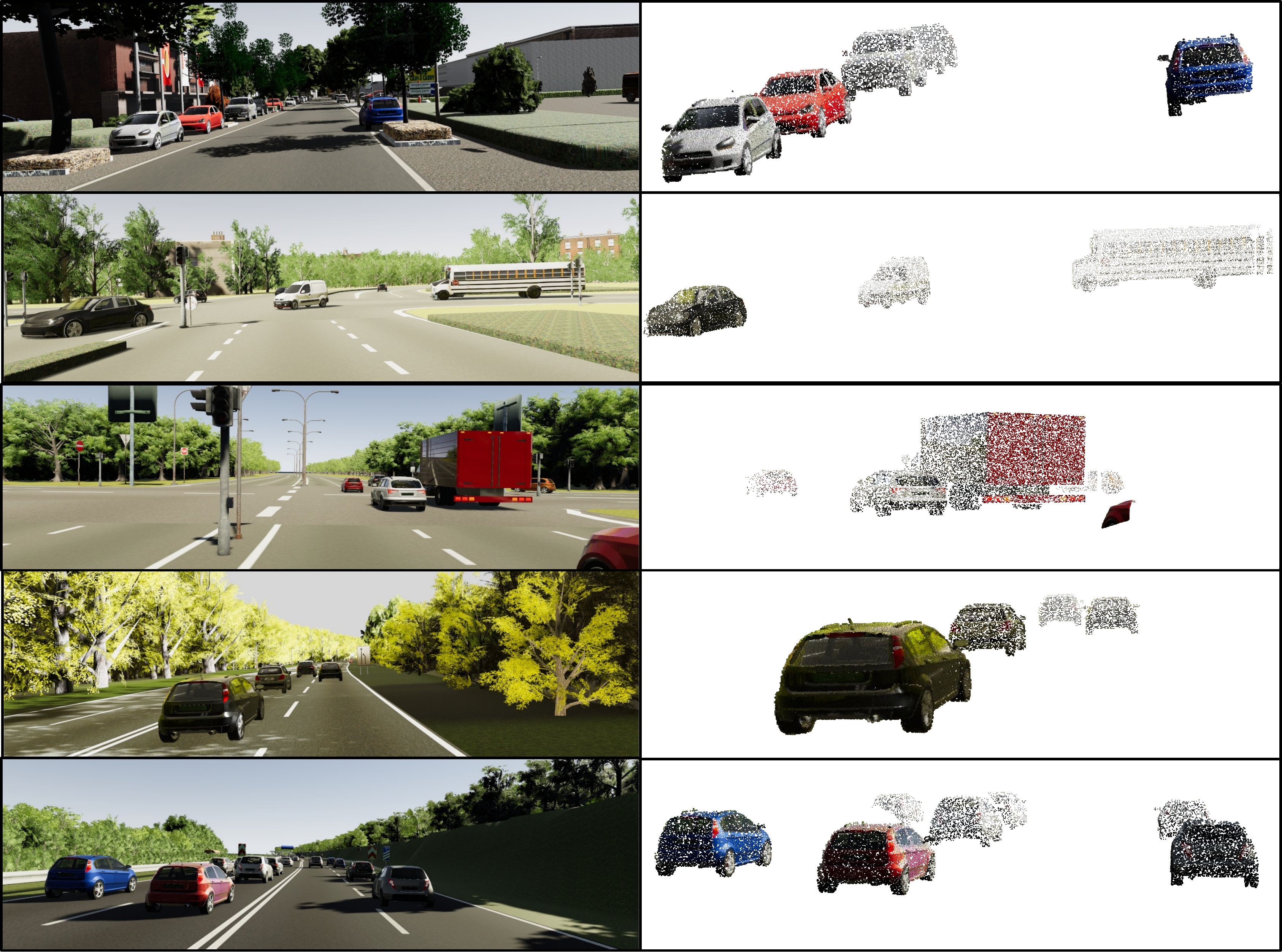}
\caption{\textbf{Sample snapshot frames from the VKS dataset} (sequence 1, 2, 6, 18, 20 from top to bottom). We consider vehicles as objects of interests as they are the major dynamic objects in a traffic scene. (Left) Original image frames from the Virtual KITTI dataset. (Right) Our created point clouds. Best viewed in color.}
\label{fig:vks}
\end{figure}
As shown in  Fig. \ref{fig:vks}, the original virtual KITTI contains five scenes of crowded
urban area (Scene01), busy intersections (Scene02, Scene06), long road in the forest (Scene18), and highway driving scene (Scene20). For each sequence, we repeatedly sample consecutive frames with a length of 10 and select the starting frame number every five frames. For all videos, we use their first $60\%$ of frames for training and the remaining $40\%$ for testing. We sample $2,048$ points for each frame and only consider points with a depth less than 35 meters.

Our VKS dataset is used for evaluating both SSFE and SPF tasks:

\begin{itemize}
    \item \textbf{SSFE}. The first five frames are used as the input. Models estimate flows from frame two to frame five. 
    \item \textbf{SPF}. Models predict the future movement of points starting from the last input frame (frame five) for five steps. Similar to our SFT3D dataset, occluded points are not considered during the evaluation.
\end{itemize}

 \subsection{Sequential Argoverse (SAG) Dataset}

 The Argoverse dataset \citep{chang2019argoverse} is collected in Pittsburgh, Pennsylvania, USA and Miami, Florida, USA by a fleet of autonomous vehicles. The collected dataset captures different seasons, weather conditions, and times of the day. We use the raw LiDAR data from \textit{Argoverse-Tracking} consisting of 113 log segments varying in length from 15 to 30 seconds. Among them, 89 logs are used for training and the rest is for testing. 
\
 \begin{figure}
    \centering
    \includegraphics[height=8cm,width=0.48\textwidth]{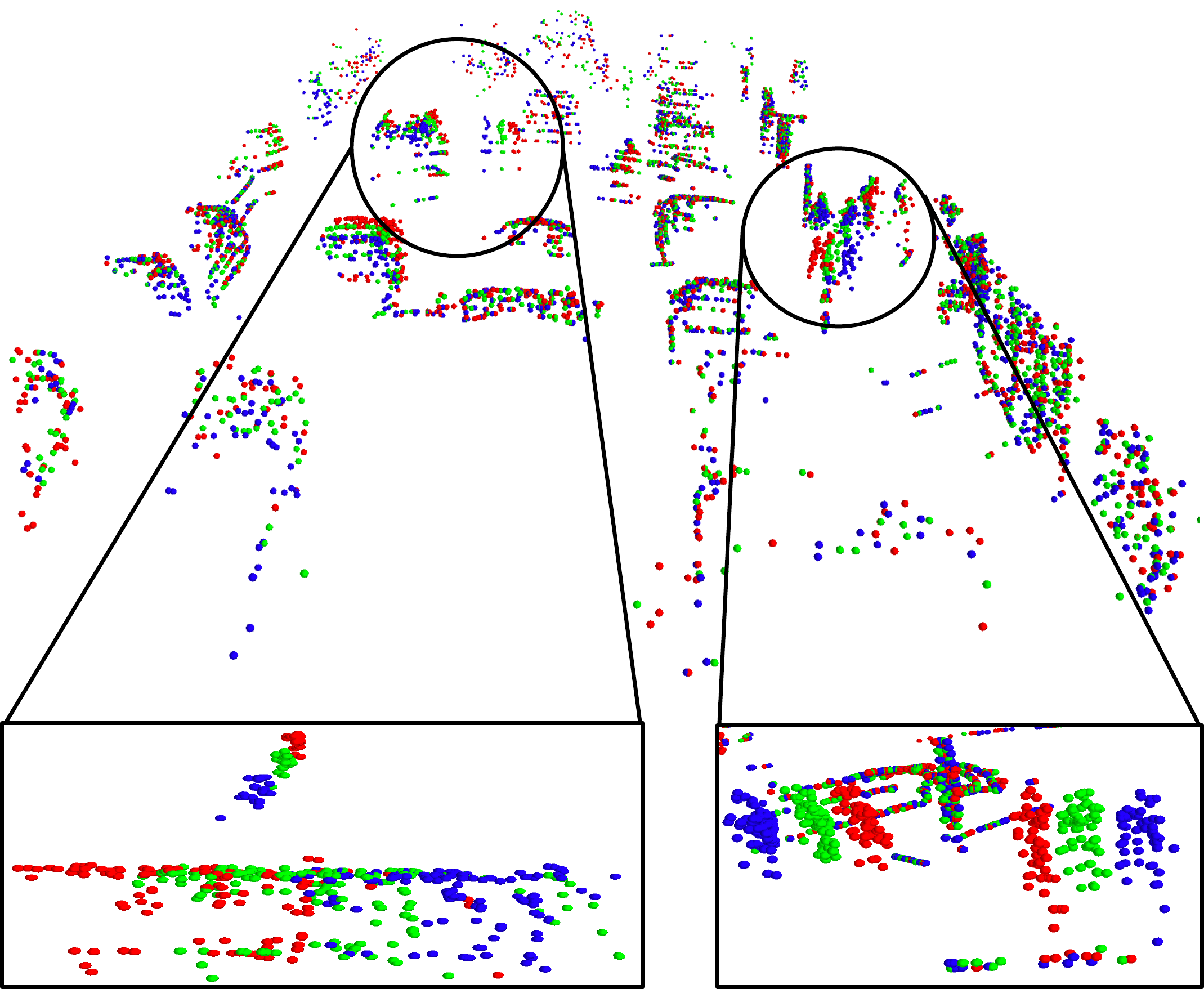}
    \caption{\textbf{Visualization of one example from our created SAG  dataset}. Different colors denote points from a different frame number: \textcolor{red}{first input frame $1$}, \textcolor{green}{last input frame $5$}, and \textcolor{blue}{last future frame $10$}. Ground plane points are removed using a heuristic algorithm. Best viewed in color.}
    \label{fig:sag}
\end{figure}

The Argoverse dataset does not provide ground-truth point-wise motion and therefore we adopt metrics that do not require annotations (Section~\ref{sec:metrics}). Hence, we only train and evaluate models on self-supervised SPF with this dataset.
 Forecasting future point clouds on real-world datasets is challenging with rapid changes in the vehicle's surroundings. We focus on short-term prediction in this work. We repeatedly sample from each driving log by randomly choosing 10 consecutive frames and creating the corresponding point cloud sequences. To achieve a reasonable computation, we sample a fixed number of points for each frame, i.e., $2,048$. We remove the ground points to reduce the bias caused by the flattened geometry of the ground. Similarly, we follow the practice of \citep{Wu_2020_CVPR} and crop the point clouds to extract region defined by  $[-32, 32] \times [-8, 8] \times [-\infty, 2]$ meters, which corresponds to the XYZ range. A sample data is shown in Fig.~\ref{fig:sag}.

\subsection{Overview of Dataset Splits}

Here we provide detailed information on the dataset splits (see Table \ref{tab:summary} as well).

\begin{itemize}
    \item \textbf{SFT3D}. A total of $5,337$ sequences are created. Among them, $4,020$ videos are used for training and the rest of $446$ and $871$ videos are held out for validation and testing, respectively. The validation set is utilized to select the best training model. 
    \item \textbf{VKS}. We have collected a total of $3,350$ point cloud sequences, where $2,175$ of them are used to train the model and the rest of the sequences are for testing.
    \item \textbf{SAG}. The SAG dataset contains a total of 1,200 test point cloud sequences, where the first five frames are the input and the rest are the ground truth future frames. Training samples are generated on-the-fly similar to test sequences.
\end{itemize}

\subsection{Evaluation Metrics} 
\label{sec:metrics}
To adapt several standard evaluation metrics in point cloud processing to fit our task format, we analyzed the usefulness of scene flow estimation metrics as well as metrics for future point cloud prediction and identified which are best suited. 
Our evaluation is mainly divided into three types: supervised  SSFE metrics, supervised SPF metrics, and self-supervised SPF metrics.

 \textbf{Supervised SSFE metrics}. If the ground truth scene flow annotations are available, we adapt the evaluation protocol of 3D scene flow estimation \citep{liu2019flownet3d,gu2019hplflownet,liu2019meteornet} and extend to sequences. Specifically, the 3D end point error (EPE3D) and accuracy (ACC3D) are used as the metrics. The
EPE3D measures the average $\ell_2$ distance between the predicted scene flow vector $\widehat{\boldsymbol{s}}_{t}^i$ and ground truth scene flow vector $\boldsymbol{s}_{t}^i$ for all points in the sequence of length $T-1$, which is computed as
\begin{equation}
    \text{EPE3D} = \frac{1}{\sum_{t,i} m_{t}^i} \sum_{t=2}^T  \sum_{i=1}^N m_{t}^i ||\widehat{\boldsymbol{s}}_{t}^i - \boldsymbol{s}_{t}^i||_2 \label{eq:supervised_distance}
\end{equation}
\noindent where $m$ is a binary mask and $m_{t}^i=0$ denotes an invalid scene flow of the point $\boldsymbol{p}_{t}^i$. This is possible in reality due to the viewpoint shift and occlusion. Taking the average over all valid points reflects the overall performance of flow estimation over sequences. 
The ACC3D reflects the portion of estimated flows that are below a specified end point error threshold among all points. Following \citep{gu2019hplflownet}, both strict and relaxed ACC3D are used:
\begin{itemize}
    \item The \textit{strict ACC3D (Acc3DS)} considers the percentage of points whose $EPE3D<0.05m$ or relative error $<5\%$.
    \item The \textit{relaxed ACC3D  (Acc3DR)} considers the percentage of points whose $EPE3D<0.1m$ or relative error $<10\%$. 
\end{itemize} 
To measure outlier prediction, we use the \textit{Outliers3D} to compute the percentage of points whose $EPE3D>0.3m$ or relative error $>10\%$. We noticed that Outliers3D is invalid when the ground truth flows are near \textbf{0}. The reason is that computing this value requires dividing by the norm of the ground truth flow, which is sensitive to values near zero.  Therefore, we fix it by proposing \textit{a rectified version of Outliers3D (RectOutliers3D)} that only depends on the condition $EPE3D>0.3m$ when the ground truth scene flow is small (e.g., its $\ell_2$ norm is lower than a threshold value 0.1). Otherwise, it remains the same to Outliers3D.

We use two additional metrics that involve projecting point clouds back to the image plane. We obtain EPE2D by computing the 2D end point error in the image plane and Acc2D by calculating the percentage of points whose EPE2D $< 3px$ or relative error $< 5\%$. 

We highlight that although all the SSFE metrics are SFE metrics that have been extended to sequences,  we maintain the same names for simplicity.

\textbf{Supervised SPF metrics}. We propose to use the standard evaluation from the trajectory forecasting community \citep{alahi2016social}. We consider two common metrics: the average displacement error (ADE) and final displacement error (FDE). 

The ADE measures the average Euclidean distance between the estimated point $\boldsymbol{p}_{T+t}^i$ and ground truth point $\boldsymbol{g}_{T+t}^i$ for all points in each prediction step. Considering the occluded points, it is defined as
\begin{equation}
    \text{ADE} =  \frac{1}{\sum_{t,i} m_{T+t}^i} \sum_{t=1}^K \sum_{i=1}^N   m_{T+t}^i ||\boldsymbol{p}_{T+t}^i - \boldsymbol{g}_{T+t}^i||_2 
\end{equation} where  $m_{T+t}^i$ denotes that the point has disappeared due to occlusion or view shift if the value is $0$, otherwise $1$. The FDE computes the average Euclidean distance between estimated point $\boldsymbol{p}_{T+K}^i$ and ground truth point $\boldsymbol{g}_{T+K}^i$ for all points at end of the prediction period $K$.

\textbf{Self-supervised SPF metrics}.  ADE and FDE are suitable when ground truth annotations are available. In most real-world datasets where ground truth point correspondences between frames are difficult to obtain, we cannot compute them anymore. We could have approximated the true correspondence by solving a weighted bipartite matching problem \citep{jonker1987shortest} or via softassign \citep{chui2000new,chui2003new} but instead adopted a simpler nearest neighbor approach \citep{barrow1977parametric,besl1992method,li2018pointcnn,fan2019pointrnn,Weng2020_SPF2}. Specifically, we generate the pseudo-ground truth points by considering nearest points to predicted points from the point cloud frame in the next timestep, and vice versa. This could be implemented as a Chamfer distance, which is previously defined in Section \ref{sec:method}, Equation \ref{eq:chamfer_distance}. We apply Chamfer distance to all future frames and sum the errors up by enumerating $t$ from $T+1$ to $T+K$. 

Similarly, we also use Earth Mover’s Distance (EMD) \citep{rubner2000earth} defined as
\begin{equation}\label{eq:emd}
\footnotesize
\mathcal{D}_{EMD}(\boldsymbol{P}_t, \widehat{\boldsymbol{P}}_t) = \min_{\phi: \boldsymbol{P}_t\rightarrow \widehat{\boldsymbol{P}}_t} \sum_{\boldsymbol{p} \in \boldsymbol{P}_t} \|\boldsymbol{p} - \phi(\boldsymbol{p})\|^2,
\end{equation} where $\phi: \boldsymbol{P}_t \rightarrow \widehat{\boldsymbol{P}}_t$ is a bijection. We divide both EMD and CD by the total number of points.

Both CD and EMD lack a mechanism to handle outliers which are likely to exist due to occlusions, noise, and sampling patterns in LiDAR point clouds. For example, in CD, noisy points or isolated points that are far from the others might substantially increase the CD by introducing large distance values between them and their nearest-neighbours, leading to noisy evaluation. In EMD, the constraint of one-to-one mapping (a bijective mapping) is usually too harsh for LiDAR point clouds, which are sampled randomly.

Inspired by \citep{yew2020rpm,gojcic2021weakly}, we introduce an evaluation method built upon optimal transport \citep{peyre2019computational}. Our goal is to find the corresponding point of the estimated point $\boldsymbol{p}_{t}^i$ with respect to the ground truth point cloud $\widehat{\boldsymbol{P}}_t$. Denoting  $\varphi$ as the hard correspondence mapping, we define an evaluation metric as
\begin{equation}\label{eq:corr}
\footnotesize
\mathcal{D}_{CORR}(\boldsymbol{P}_t, \widehat{\boldsymbol{P}}_t) = \sum_{\boldsymbol{p} \in \boldsymbol{P}_t} \|\boldsymbol{p} - \varphi(\boldsymbol{p}, \widehat{\boldsymbol{P}}_t)\|^2.
\end{equation} 

To obtain $\varphi$, we firstly obtain the optimal soft assignment (called the softassign in an homage to the softmax nonlinearity) $\varphi_{soft}$ via the Sinkhorn algorithm \citep{sinkhorn1964relationship}, and then follow \citep{li2021self} to obtain hard correspondence. Specifically, we use the point coordinates of $\boldsymbol{p}_t^i$ and $\widehat{\boldsymbol{p}}_t^j$ to construct an affinity matrix $\mathcal{M}$ where  $\mathcal{M}_{\boldsymbol{p}_t^i, \widehat{\boldsymbol{p}}_t^j}$
is defined as
\begin{equation}
    \mathcal{M}_{\boldsymbol{p}_t^i, \widehat{\boldsymbol{p}}_t^j} = - exp(\frac{\gamma}{ d(\boldsymbol{p}_t^i, \widehat{\boldsymbol{p}}_t^j) + \epsilon}).
\end{equation}

Here, $\gamma$ is empirically set to $10$, $\epsilon = 1e^{-8}$, and $d(\boldsymbol{p}_t^i, \widehat{\boldsymbol{p}}_t^j)$ is the Euclidean distance. Given $\mathcal{M}$, we perform an alternating row and column normalization for a few iterations (e.g., five iterations), which yields a doubly stochastic assignment matrix $\mathcal{A}$. The soft correspondence function $\varphi_{soft}$ then reads $\varphi_{soft}(\boldsymbol{p}_{t}^i, \widehat{\boldsymbol{P}}_t) = a_i/|a_i|_1 \widehat{\boldsymbol{P}}_t$ where $a_i$
is the i-th row
of $\mathcal{A}$.

To handle outliers, we follow \citep{chui2003new,yew2020rpm,gojcic2021weakly} and add an additional row and column of ones (a slack row and column) to the original input of Sinkhorn normalization (the matrix $\mathcal{M}$) while only performing the alternating row and column normalization on non-slack rows and columns to obtain a resulting matrix $\hat{\mathcal{A}}$ . To generate $\varphi$, we follow \citep{li2021self} to modify each row of $\hat{\mathcal{A}}$ by setting its column element with the maximum value to 1 and the remaining element to 0 such that the point with the highest transport score is selected as the corresponding point in this row. 

We average all the $\ell_2$ distances between all pairs of points except those which are assigned to slack columns.  These are denoted as $\boldsymbol{P}_t^{valid} \subset \boldsymbol{P}_t$ respectively and then the Sinkhorn Distance (SD) evaluation metric is defined as 
\begin{equation}\label{eq:sd}
\footnotesize
\mathcal{D}_{SD}(\boldsymbol{P}_t, \widehat{\boldsymbol{P}}_t) = \frac{1}{|\boldsymbol{P}_t^{valid}|}\sum_{\boldsymbol{p} \in \boldsymbol{P}_t^{valid}} \|\boldsymbol{p} - \varphi(\boldsymbol{p}, \widehat{\boldsymbol{P}}_t)\|^2.
\end{equation} 

We use the ADE, FDE, CD, and EMD metrics for evaluating SPF tasks on SFT3D and VKS datasets. When we move to the SAG dataset, we report both CD, EMD, and SD metrics because no ground truth annotation is available. Also, the introduced SD is used to downweight outliers. Note that SPCM-Net uses the CD as the learning objective on the SAG dataset.

\section{Experiments}\label{sec:experiment}

In this section, our main goal is to present experimental results evaluating SPCM-Net and relevant baselines on the proposed SFT3D, VKS, and SAG datasets.

To that end, we first evaluate SPCM-Net and relevant models on supervised SSFE and supervised SPF on the SFT3D dataset (Section \ref{sec:sft3d}). Then, we evaluate the two taskson the VKS dataset  with the same models  (Section \ref{sec:vks}). Next, we evaluate SPCM-Net on the self-supervised SPF task with the SAG dataset (Section \ref{sec:sag}). Finally, we demonstrate the benefit of our recurrent cost volume approach to modeling point cloud sequences in a controlled experiment using the standard KITTI scene flow dataset (Section \ref{sec:kitti}). 

\subsection{Implementation Details}
We implemented all the developed models in PyTorch \citep{paszke2019pytorch} using distributed training with 8 GPUs.  Training each model generally takes 1-2 days. For SFE models (such as FlowNet3D) built upon TensorFlow \citep{abadi2016tensorflow}, we converted their pre-trained model weights into Pytorch and achieved identical performance. For most of the experiments, we set the learning rate and weight decay as 0.001 and 0.0001, respectively.  We trained the models for 400 epochs while decaying the learning rate of each parameter group by 0.1 every 100 epochs. The gradient clip technique was applied to normalize the gradients. We didn't use any data augmentation strategy (such as rotation and scaling). We will release training and evaluation code for the new benchmark and models to facilitate future research at \url{https://github.com/BestSonny/SPCM}.

\begin{table*}[t]
    \centering
    \caption{\textbf{Multi-step representation achieves improved performance on the SSFE task for point cloud sequences longer than two frames}. We verify it by comparing supervised SSFE results against the adapted prior arts on our SFT3D validation and test splits and VKS dataset. `FT3D' denotes training with the pair-wise FlyingThings3D  dataset (FT3D) prepared by \citep{liu2019flownet3d}.}
    \normalsize
    \resizebox{0.98\textwidth}{!}{  
    \begin{tabular}{l|c|cccc|cc}
         \hline\noalign{\smallskip}
        \multicolumn{8}{c}{\textbf{SFT3D Validation Split}}
         \\ 
        \hline\noalign{\smallskip}
        Method         & EPE3D$\downarrow$           & Acc3DS$\uparrow$          & Acc3DR$\uparrow$          & Outliers3D$\downarrow$      &
        RectOutliers3D$\downarrow$      & EPE2D$\downarrow$           & Acc2D$\uparrow$           \\
        \noalign{\smallskip}\hline\hline 
        FlowNet3D~\citep{liu2019flownet3d}  + FT3D & 0.200        &  0.173        &  0.494       &  0.783      & 0.764  & 11.803     &   0.319     \\
        FLOT~\citep{puy20flot} + FT3D & 0.173 &  0.304 & 0.606  &0.649 &  0.630   & 10.145   & 0.412   \\
        PointPWC-Net~\citep{wu2019pointpwc} + FT3D &  0.190  &  0.318   & 0.615  & 0.635 & 0.625 & 11.178  & 0.415   \\
        \hline 
        \hline 
       
         FlowNet3D \citep{liu2019flownet3d} + SFT3D & 0.136        &   0.314         &  0.692          &    0.614          & 0.595 & 7.939      &    0.466      \\
        FLOT~\citep{puy20flot} + SFT3D & 0.159   & 0.331   & 0.639   &   0.619 & 0.600 & 9.513 & 0.440   \\
        PointPWC-Net \citep{wu2019pointpwc} + SFT3D & 0.115  & 0.455  &  0.760 & 0.502 & 0.483 &   7.210  &  0.548\\
        \hline 
        \hline 
         SPCM-Net (Ours)  & \textbf{0.108} &  \textbf{0.484}  &  \textbf{0.782} &  \textbf{0.468} & \textbf{0.450} & \textbf{6.709}  & \textbf{0.567}  \\
        \hline 
    
     \hline\noalign{\smallskip}
        \multicolumn{8}{c}{\textbf{SFT3D Test Split}}
         \\ 
    \hline\noalign{\smallskip}
        Method         & EPE3D$\downarrow$           & Acc3DS$\uparrow$          & Acc3DR$\uparrow$          & Outliers3D$\downarrow$      &
        RectOutliers3D$\downarrow$      & EPE2D$\downarrow$           & Acc2D$\uparrow$           \\
        \noalign{\smallskip}\hline\hline
        FlowNet3D~\citep{liu2019flownet3d} + FT3D & 0.191         & 0.169         & 0.494          & 0.792       &  0.769 & 11.743       & 0.320       \\
        FLOT~\citep{puy20flot}  + FT3D & 0.183   & 0.287  & 0.583  &  0.676 & 0.653 & 11.364   & 0.398     \\
        PointPWC-Net~\citep{wu2019pointpwc} + FT3D &  0.178   &  0.321   & 0.606   & 0.654 & 0.640 & 11.313  & 0.418     \\
        \hline 
        \hline 
        FlowNet3D \citep{liu2019flownet3d} + SFT3D & \textbf{0.150}        & 0.278         & 0.636         &  0.676          & 0.652 &  \textbf{9.327}        &  0.432       \\

        FLOT~\citep{puy20flot} + SFT3D & 0.172   &  0.310   &  0.612   & 0.651 & 0.628 &   10.763   & 0.422    \\
        PointPWC-Net \citep{wu2019pointpwc} + SFT3D  & 0.174   &  0.320   & 0.609    & 0.656   & 0.633  &  11.087   & 0.420   \\
        \hline 
        \hline 
        SPCM-Net (Ours) & 0.157  & \textbf{0.380}  & \textbf{0.659} & \textbf{0.597} & \textbf{0.575} &  10.204  & \textbf{0.473}  \\
         \hline 
         \hline\noalign{\smallskip}
        \multicolumn{8}{c}{\textbf{VKS Dataset}}
         \\ 
         \hline\noalign{\smallskip}
        Method         & EPE3D$\downarrow$           & Acc3DS$\uparrow$          & Acc3DR$\uparrow$          & Outliers3D$\downarrow$  & RectOutliers3D$\downarrow$ & EPE2D$\downarrow$           & Acc2D$\uparrow$      \\
        \noalign{\smallskip}\hline\hline
 FlowNet3D \citep{liu2019flownet3d} &  0.0584      & 0.7178      & 0.8819   & 0.4628      & 0.2622 & 2.4355   & 0.8119     \\
        FLOT~\citep{puy20flot}  & 0.0672  & 0.7622   & 0.8638 & 0.3944  & 0.2095 & 2.5508  & 0.8354 \\
        PointPWC-Net \citep{wu2019pointpwc} & 0.0458   & 0.8085   &  0.8990   & 0.3700 &  0.1793 & 1.9018  & 0.8591  \\
        \hline 
        \hline 
         SPCM-Net (Ours) & \textbf{0.0454}   &  \textbf{0.8330}   &  \textbf{0.9121}   & \textbf{0.3520}  &  \textbf{0.1634} & \textbf{1.7578}  &  \textbf{0.8836}  \\
         \hline 
         
    \end{tabular}
    }
    \label{tab:sft3d_val}
\end{table*}

\begin{table*}[t]
\centering
\caption{\textbf{SPCM-Net can be adapted to support the SPF task while achieving a superior performance. Pretraining on the SSFE task helps the SPF task}. We show supervised SPF results on our SFT3D datasets. All models are trained from scratch except the model (SPCM-Net (Ours) + Pretrained). `Pretrained' denotes that it is finetuned on the model pretrained in the task of supervised SSFE.}
\small
\resizebox{0.85\textwidth}{!}{  
\begin{tabular}{c|c|c|c|c|c|c|c|c}
\hline \noalign{\smallskip}
 \multirow{2}{*}{Method}  & \multicolumn{4}{c|}{\textbf{SFT3D Validation Split}}   & \multicolumn{4}{c}{\textbf{SFT3D Test Split}}      \\\noalign{\smallskip} \cline{2-9} \noalign{\smallskip}
                       & ADE $\downarrow$           &  FDE$\downarrow$          & CD$\downarrow$          & EMD$\downarrow$    & ADE $\downarrow$           &  FDE$\downarrow$          & CD$\downarrow$          & EMD$\downarrow$    \\ \noalign{\smallskip} \hline \hline
  PointNet++\citep{qi2017pointnet} + LSTM                       & 0.6740 & 1.0045 & 0.4603 & 0.9495 & 1.4017 & 2.1717 & 0.8878 & 2.0773 \\ 
PointRNN~\citep{fan2019pointrnn}            & 0.5605 & 0.8022 & 0.3715 & 0.7998 & 0.7607 & 1.1275 & 0.4783 & 1.0996  \\ \hline \hline
SPCM-Net (Ours)          & \textbf{0.5069} & \textbf{0.7893} & \textbf{0.3300} & \textbf{0.7364} & \textbf{0.6286} & \textbf{0.9769} & \textbf{0.3848} & \textbf{0.9155} \\ 
SPCM-Net (Ours) + Pretrained & \textbf{0.2488} & \textbf{0.3886} & \textbf{0.1737} & \textbf{0.3408} & \textbf{0.3978} & \textbf{0.6373} & \textbf{0.2514} & \textbf{0.5684} \\ \hline
\end{tabular}
}
\label{tab:sft3d_future}
\end{table*}
\begin{table*}[t]
\centering
\caption{\textbf{Our general \textit{SPCM-Net} architecture achieves competitive SPF results compared to tailored state-of-the-art models under both supervised and self-supervised settings.}}
\footnotesize
 \resizebox{0.85\textwidth}{!}{  
\begin{tabular}{c||c|c|c|c||c|c|c}
\hline \noalign{\smallskip}
 \multirow{2}{*}{Method}  & \multicolumn{4}{c||}{\textbf{VKS (supervised)}}  & \multicolumn{3}{c}{\textbf{SAG (self-supervised)}}       \\\noalign{\smallskip} \cline{2-8} \noalign{\smallskip}
                       & ADE $\downarrow$           &  FDE$\downarrow$          & CD$\downarrow$          & EMD$\downarrow$        & CD$\downarrow$          & EMD$\downarrow$ & SD$\downarrow$    \\ \noalign{\smallskip} \hline \hline
 PointNet++~\citep{qi2017pointnet++} + LSTM  & 1.1747  & 1.9278 & 0.7260 & 1.4071 & 2.0718 & 2.5574  & 2.4363 \\
 PointRNN~\citep{fan2019pointrnn}& 0.2856  & \textbf{0.4655}  & 0.1551 & 0.3575 & \textbf{1.2322} & 2.3160  & \textbf{1.3630}\\   \hline \hline
SPCM-Net (Ours) & \textbf{0.2768}  & 0.4799  & \textbf{0.1400}  & \textbf{0.3418}  & 1.3453 & \textbf{2.2992} & 1.4845 \\ \hline
\end{tabular}
}
\label{tab:vks_future}
\end{table*}
\begin{figure*}[t]
    \centering
    \includegraphics[width=\textwidth]{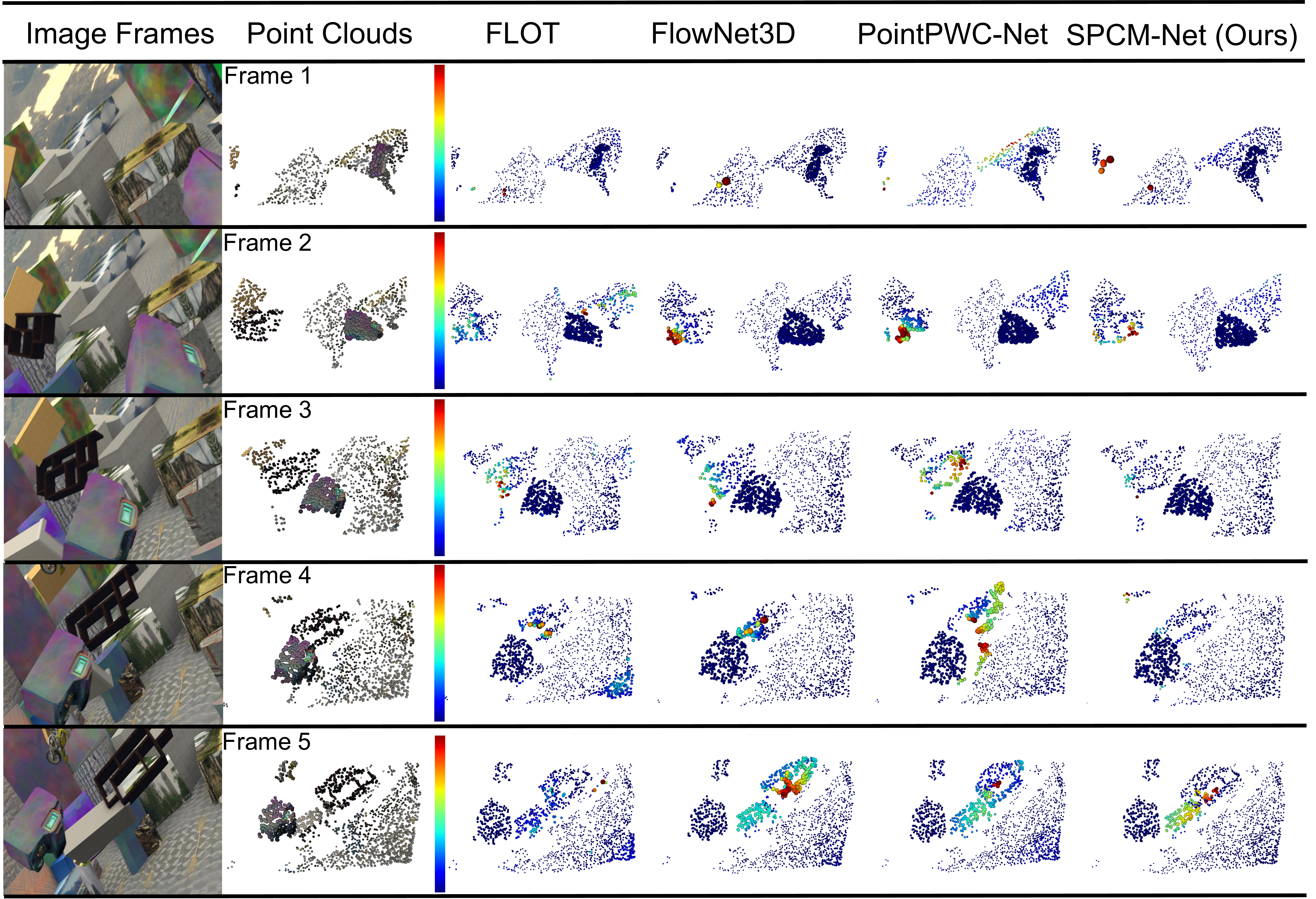}
    \caption{\textbf{The SPCM-Net can generate accurate scene flow estimation while largely reducing the outlier prediction, due to its capability of extracting multi-step information from sequences. Points can  leverage their historical motion patterns in nearby locations to estimate more accurate and robust scene flow.} Qualitative comparisons between SSFE results of different models on our SFT3D dataset. The `Image Frames' column visualizes one sample sequence of the original FlyingThings3D dataset. The `Point Clouds' column shows the corresponding point cloud sequences created after adding ground truth scene flows to themselves. All the models are trained with our SFT3D dataset. The error heatmaps illustrate the estimation results. The errors gradually increase from dark blue to dark red.  Best viewed in color.}
    \label{fig:sft3d_visual}
\end{figure*}
\subsection{Sequential FlyingThings3D Dataset (SFT3D)}\label{sec:sft3d}
 
\subsubsection{Supervised SSFE} 
\textbf{Models}. We created several models to compare against the proposed SPCM-Net by directly adapting prior SFE approaches for frame pairs. We selected three representative state-of-the-art architectures that are publicly available for a comprehensive evaluation, namely FlowNet3D \citep{liu2019flownet3d}, PointPWC-Net \citep{wu2019pointpwc}, and FLOT \citep{puy20flot}. Originally, these models only support scene flow estimation between two consecutive frames. FlowNet3D and FLOT were trained with the FlyingThings3D  dataset (FT3D) prepared by \citep{liu2019flownet3d}. We train PointPWC-Net using the same dataset. These models are denoted as $\textbf{MODEL\_NAME + FT3D}$.

To report the performance for these models on our new SFT3D dataset, we pass every two consecutive frames of each point cloud sequence to obtain the predicted scene flows (e.g., four-step scene flow estimation for a point cloud sequence of length five). To ensure a fair comparison, we customized these methods to support SSFE by making n-step predictions and retraining them with the training split of the SFT3D dataset. We used the validation split to select the best models. These models are  denoted as $\textbf{MODEL\_NAME + SFT3D}$. The same setting of training and evaluation ensures a fair comparison between SPCM-Net and these models.

\textbf{Results.}  All results are aggregated and shown in Table~\ref{tab:sft3d_val}.  All models trained with the pair-wise FT3D dataset achieve limited performance on the SFT3D dataset. After customizing these models and re-training them on SFT3D, we observe consistent improvement in performance. The improvement reflects that the extra supervision from multiple frame pairs of a sequence provides a stronger learning signal compared to single-frame-pair supervision. However, these models independently conduct standard scene flow estimation on all frame pairs in point cloud sequences, ignoring multi-step spatiotemporal information. This has been addressed by SPCM-Net, which can recurrently process point cloud sequences.

\begin{figure*}[t]
    \centering
     \includegraphics[width=\textwidth]{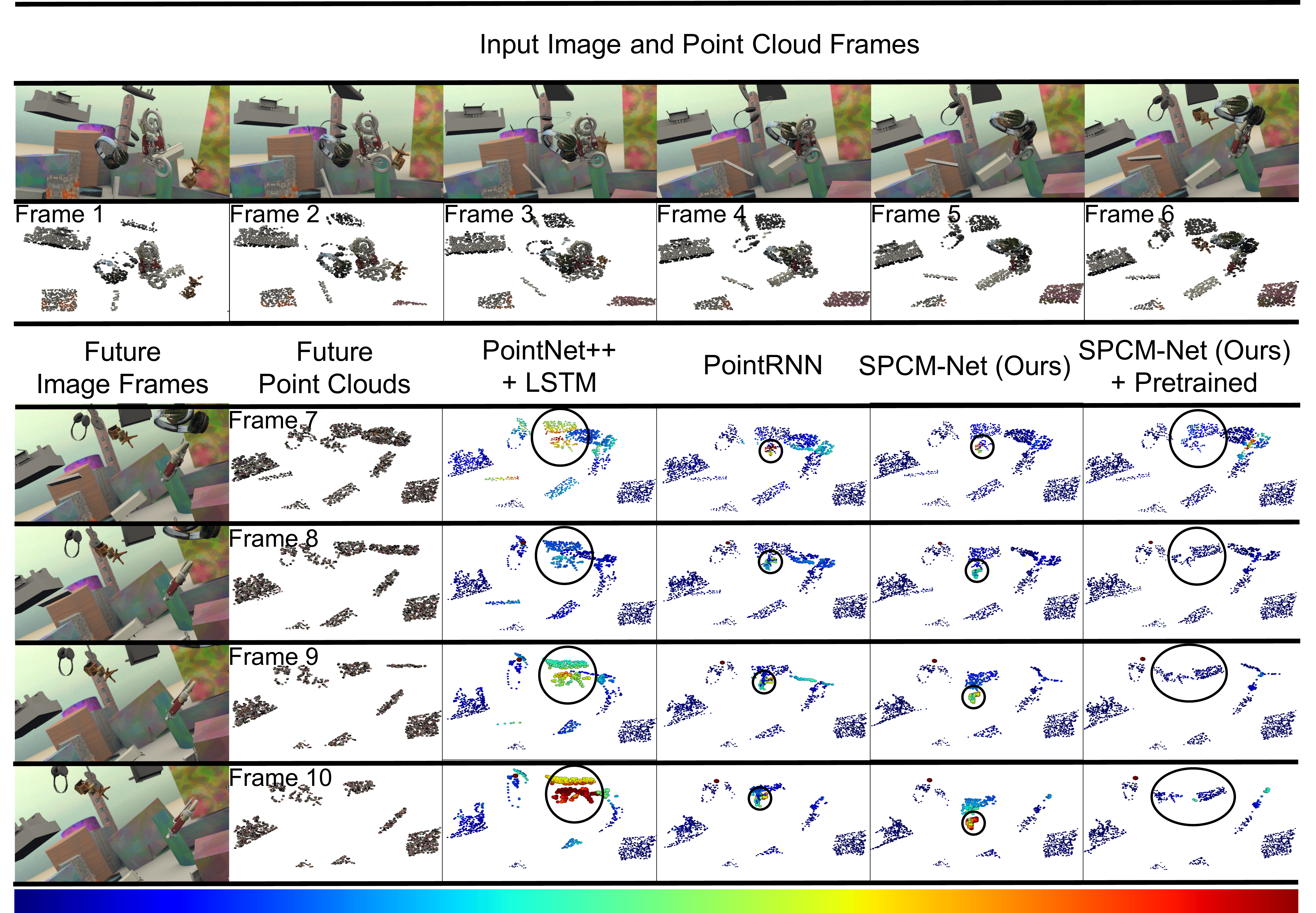}

    \caption{Qualitative comparisons between SPF results of different models on our SFT3D dataset. Top rows show the original input image frames and our constructed point cloud frames. The bottom rows show the prediction results. The error heatmaps reflect the average displacement errors (ADE) of points. The errors gradually increase from dark blue to dark red. \textbf{We  show  that  performing  pre-training  on  SSFE, followed by fine-tuning on the SPF task, helps boost the performance of SPF compared to training from scratch}, which is verified by the circled results. `SPCM-Net (ours) + Pretrained' successfully predicts the future motion of the lamp with the wooden mount while all models trained from scratch have failed. Best viewed in color.}
    \label{fig:sft3d_future_visual}
\end{figure*}
SPCM-Net surpasses baseline methods significantly on various metrics. On both validation and test splits of the SFT3D dataset, SPCM-Net shows a clear improvement over the best models. It obtains EPE3D scores of $0.108$ and $0.157$ on the validation and test split, respectively,  improving the best results of relevant models, i.e., \textbf{PointPWC-Net + SFT3D} and \textbf{FlowNet3D + SFT3D}. The improvements become larger when we measure the accuracy of scene flow prediction and the number of outlier predictions. For example, on the test split, SPCM-Net achieves a Acc3DS score of $0.380$ and a Acc3DR score of $0.659$, largely outperforming FlowNet3D, FLOT, and PointPWC-Net.

Specifically, it surpasses the previous best result on every single metric: $>6\%$ on Acc3DS (higher is better), $>2\%$ on Acc3DR (higher is better), and $>5\%$ on Outliers3D (lower is better). This indicates that our framework can handle sequences in a more principled way by recurrently processing a point cloud sequence, thus capturing scene dynamics across a longer temporal length. Compared to baselines designed for pair-wise frames, the capability to utilize multi-step information reduces outlier predictions.
Overall, our model demonstrates an ability to utilize historical point-wise motion patterns derived from spatiotemporal neighborhoods of points across frames. We further verify this by visualizing model predictions on the SFT3D dataset, as shown in  Fig. \ref{fig:sft3d_visual}.

\subsubsection{Supervised SPF}

\noindent \textbf{Models}: We evaluate the prediction task by comparing against several prediction baselines: 1)  \textbf{PointNet++} \citep{qi2017pointnet++} \textbf{+ LSTM}. We established a simple baseline by converting each point cloud into a global feature vector via a pooling layer. To learn the temporal dynamics and propagate it to the future, we use a standard fully-connected LSTM network to process the global feature vectors of the past input frames. At each timestep, the output feature of the LSTM will be broadcasted to each point and combined with the local point feature similar to the segmentation network in PointNet++. The model output will be the motion offsets of future points. 2) We used a recent preprint work  called \textbf{PointRNN} \citep{fan2019pointrnn}, which essentially extends the \textit{flow embedding} layer in FlowNet3D \citep{liu2019flownet3d} to a recurrent model to support future prediction.
We are unable to include the SPFNet architecture~\citep{Weng2020_SPF2} in our evaluation as code has not been released for it at the current time, but we aim to add it to our benchmark in the near future. 

We evaluate two variants of the proposed SPCM-Net---one trained from scratch (\textbf{SPCM-Net}) and one fine-tuned on SPF after pre-training on SSFE (\textbf{SPCM-Net + Pretrained}). The first one ensures a fair comparison to the baselines while the second one  explores whether pretraining on the SSFE task helps the SPF task. All the models were trained with our proposed SFT3D dataset. 

\textbf{Results.} Table \ref{tab:sft3d_future} reports the future prediction results on validation and test splits. Compared to other baseline approaches, our SPCM-Net  achieves lower  $\text{ADE}$, $\text{FDE}$, $\text{CD}$, and $\text{EMD}$ under the same setting of training from scratch. Additionally, we verify that using the pre-trained weights obtained from the SSFE task to initialize the model is beneficial. It further reduces the displacement errors and obtains better future predictions. For example, on both validation and testing splits, compared to the best result of SPCM-Net, SPCM-Net + Pretrained further reduces the ADE and FDE to $0.2488$ and $0.3978$, achieving a significant decrease of $50.92\%$ and $36.72\%$, respectively. The $\text{CD}$ and $\text{EMD}$ also reflect that a significant improvement has been achieved. 

We visualize a sample generated by each model in Fig. \ref{fig:sft3d_future_visual}. We see that SPCM-Net + Pretrained achieves the best future prediction. In particular, it successfully predicts the motion of the lamp with the wooden mount while all other models have failed.

\subsection{Virtual KITTI Sequence (VKS) Dataset}\label{sec:vks}

The VKS dataset further allows us to evaluate models on simulated environments for the real-world, providing a preliminary prototype evaluation for traffic scenes. It provides accurate ground truth scene flows and future movements of multiple moving objects that are rather difficult to obtain in real-world settings.
\subsubsection{Supervised SSFE}

\begin{figure*}[th!]
    \centering
    \includegraphics[height=7cm, width=0.85\textwidth]{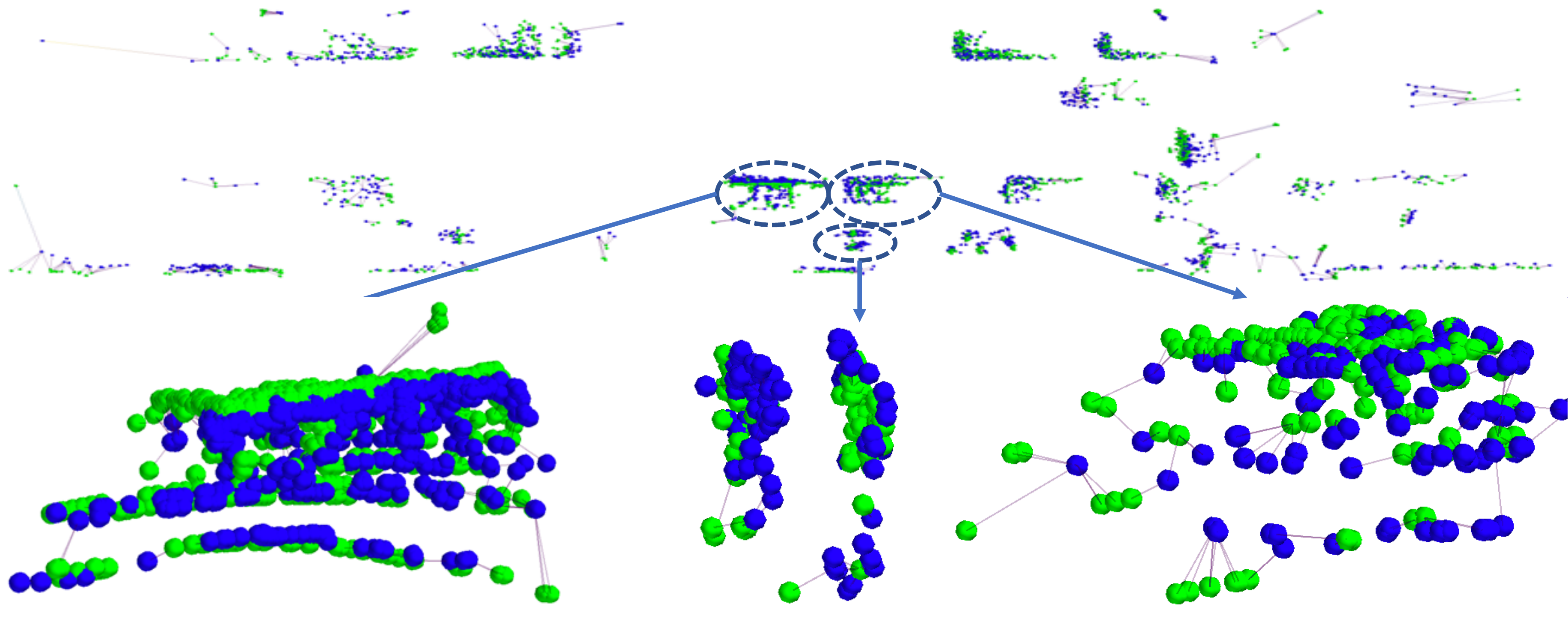}
    \caption{Visualization of SAG future prediction with our SPCM-Net. \textbf{Top}: Top view of the whole scene. \textbf{Bottom}: Zoomed in areas. \textcolor{green}{green} points are future points at frame $10$ while \textcolor{blue}{blue} points are the predicted point cloud. Lines connect the nearest neighboring points between \textcolor{green}{green} points and \textcolor{blue}{blue} points, which are obtained by the Chamfer distance. Shorter-length lines are preferred since they reflect lower errors.}
    \label{fig:sag_vis}
\end{figure*}
\begin{figure}[t]
    \centering
    \includegraphics[width=0.49\textwidth]{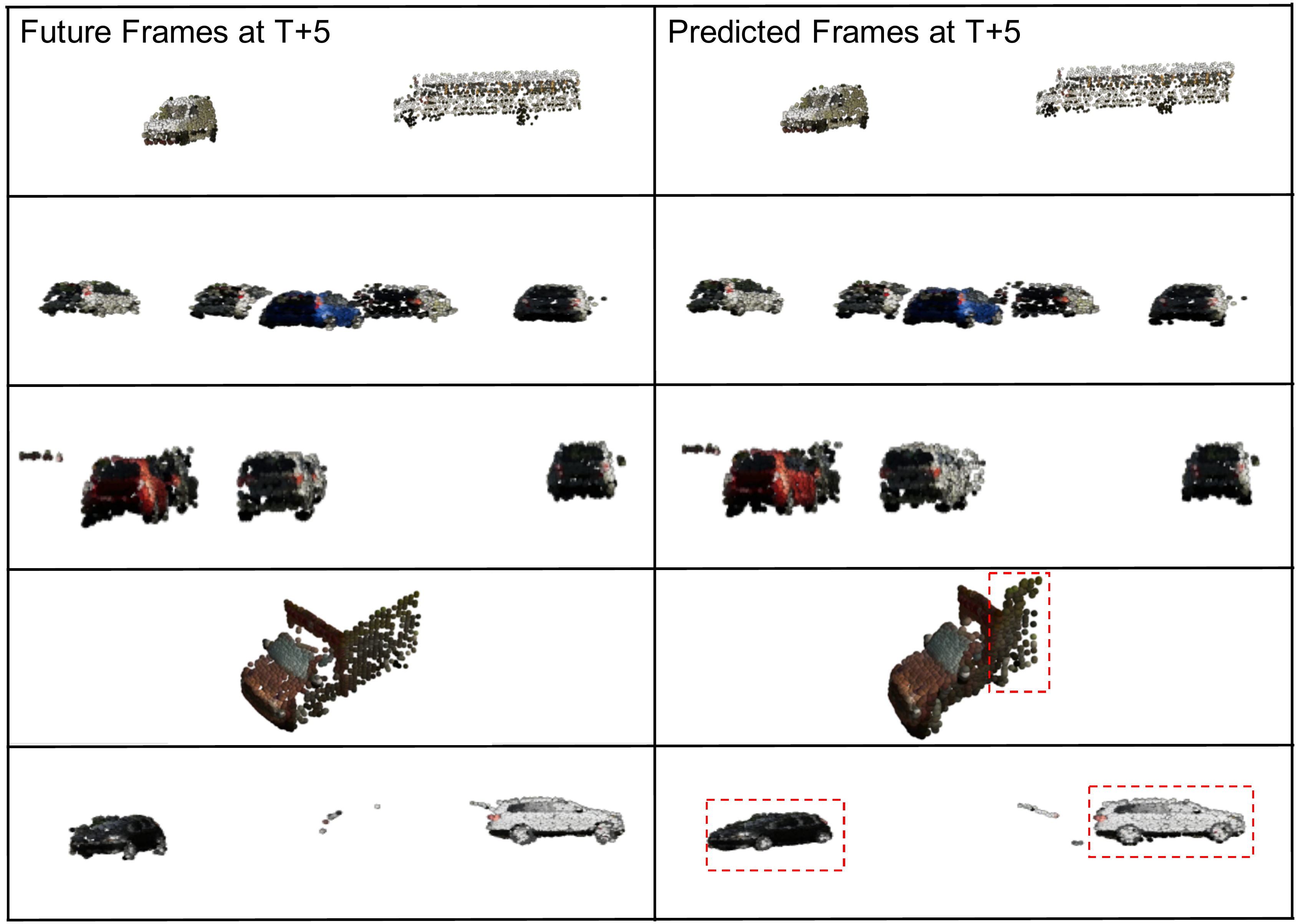}
    \caption{Visualization of VKS future prediction with our SPCM-Net. Left: the ground truth future frames at the end of prediction period (the $5^\text{th}$ future frames). Right: the predicted frames. Best viewed in color.}
    \label{fig:vks_visual}
\end{figure}

\textbf{Baselines.} The evaluated models are very similar to the models used in the SFT3D dataset except for one difference. Both PointPWC-Net and SPCM-Net have changed to ball-query-based neighbor search instead of K-nearest neighbor search for better performance. This follows the practice of Qi et al. \citep{qi2017pointnet++} to maintain a fixed region scale and highlight local region patterns. All models are first pre-trained on the SFT3D dataset then fine-tuned on the VKS dataset.

\textbf{Results.} The results are shown in Table \ref{tab:sft3d_val}. We find that SPCM-Net outperforms other baseline methods on all the metrics including EPE3D, Acc3DS, Acc3DR, Outliers3D, RectOutliers3D, and EPE2D, and Acc2D. The main improvements are seen in the accuracy scores (reflected by Acc3DS and Acc3DR) and reductions in outlier predictions. FLOT  \citep{puy20flot} performs slightly worse than we initially expected. We suspect that the K-nearest neighbor search used in FLOT could potentially degrade the performance, due to its learned features that are less generalized in space.

\subsubsection{Supervised SPF}
\textbf{Models.} The baselines are the same as for the SFT3D dataset. Our SPCM-Net uses a ball-query-based neighbor search. All models are first pre-trained on the supervised SPF task with the SFT3D dataset and then fine-tuned on the VKS dataset.

\textbf{Results.} Quantitative results are listed in Table \ref{tab:vks_future}. We show that SPCM-Net achieves a  comparable performance compared to  PointRNN~\citep{fan2019pointrnn}. Interestingly, PointNet++~\citep{qi2017pointnet++} + LSTM performs significantly worse. Because it uses the global fully-connected feature as the spatiotemporal representation, it struggles to capture the dynamics of local regions in the VKS dataset.

A visualization of predictions made by our SPCM-Net is provided in Fig. \ref{fig:vks_visual}. It makes reasonably good predictions, and overall we achieve competitive performance to the prior art PointRNN.
We show some typical errors made by SPCM-Net. When predicting the future movement of the truck objects, the model shows that it lacks awareness of the physical constraints.  Another challenge is that when a vehicle makes a turn, the model struggles to capture the precise movement (i.e., orientation, speed) of the vehicle. This could be found in the black car example in Fig. \ref{fig:vks_visual}. We encourage the community to further explore this problem by investigating advanced topics in generative models \citep{achlioptas2018learning,wang2020sceneformer} and physical scene understanding \citep{yao20183d}.

\subsection{Sequential Argoverse (SAG) Dataset}\label{sec:sag}

To explore the performance in real-world datasets, we train and evaluate models on the SAG dataset. The baselines are the same as models used in the supervised SPF VKS experiment, except that we use the self-supervised SPF objective to guide model learning.

\textbf{Results}. Results are shown in the right section of Table \ref{tab:vks_future}. SPCM-Net achieves a competitive result compared to other baselines. Fig. \ref{fig:sag_vis} shows visual results. The overall prediction is relatively good with high fidelity. However, we notice that the ground truth points formulated by randomly sampling points in a scene contain outliers (lines of long length in Fig. \ref{fig:sag_vis}). This can explain high CD and EMD scores. Both PointRNN and SPCM-Net achieve competitive SD results. We encourage future work to revisit the point sampling process by focusing on object points. However, this likely requires applying extra unsupervised object segmentation  \citep{landrieu2018large} or object discovery techniques \citep{karpathy2013object}. Also, it is promising to map point clouds to a latent space where a robust distance metric can be computed, as evidenced in a recent work \citep{zuanazzi2020adversarial} using adversarial learning.

\subsection{KITTI Scene Flow Dataset}\label{sec:kitti}
We conduct an experiment on standard scene flow estimation using the KITTI Scene Flow benchmark with the multi-frame setup of MeteorNet~\citep{liu2019meteornet} to demonstrate a direct comparison between their architecture and SPCM-Net.
We consider MeteorNet as the 4D extension of PointNet++ \citep{qi2017pointnet++} since it appends a 1D temporal coordinate to the 3D spatial coordinates. The resulting 4D coordinates help find spatiotemporal neighbors and extract features in 4D space to handle point cloud sequences.

\textbf{Experimental setup}. The KITTI scene flow dataset \citep{menze2015object} provides ground truth disparity maps and optical flows for 200 frame pairs, from where the 3D ground truth scene flow can be constructed. 
Among $200$ frame pairs, only $142$ provides the corresponding mapping to laser point clouds. MeteorNet \citep{liu2019meteornet} further extends the dataset to use preceding point cloud frames. As a result, the task aims to predict one-step scene flow of each frame pair while taking a point cloud sequence as the input (recall from Fig.~\ref{fig:problem}b). The first $100$ of the $142$ frames are used to fine-tune the models while the remaining $42$ sequences are used for testing. We used the same dataset prepared and released by \citep{liu2019meteornet} for training and evaluation.

 \begin{table}
\footnotesize
\centering
\caption{\textbf{Flow estimation results on KITTI sceneflow dataset.} Metrics are the mean and standard deviation of the End-point-error (EPE) of scene flow.}
\begin{tabular}{l|c|cc}

\hline\noalign{\smallskip}
Method    & Frames & Mean & Std  \\ \noalign{\smallskip}\hline\noalign{\smallskip} 
FlowNet3D   & 2 & 0.287  & 0.250 \\
MeteorNet (direct)    & 3  & 0.282 & 0.204 \\
MeteorNet (direct)  & 4  & 0.263 & 0.210 \\ 
MeteorNet (chained-flow)     & 3  & 0.277 & 0.244   \\
MeteorNet (chained-flow)   & 4  & 0.251 & 0.227  \\ \noalign{\smallskip} \hline \hline\noalign{\smallskip} 
SPCM-Net (ours)  & 3  & \textbf{0.229} &  \textbf{0.184} \\
SPCM-Net (ours)  & 4  & \textbf{0.194} & \textbf{0.174} \\\noalign{\smallskip} \hline
\end{tabular}
\label{tab:kitti_flow}
\end{table}

 \begin{figure}[th!]
    \centering
    \includegraphics[height=10cm,width=0.485\textwidth]{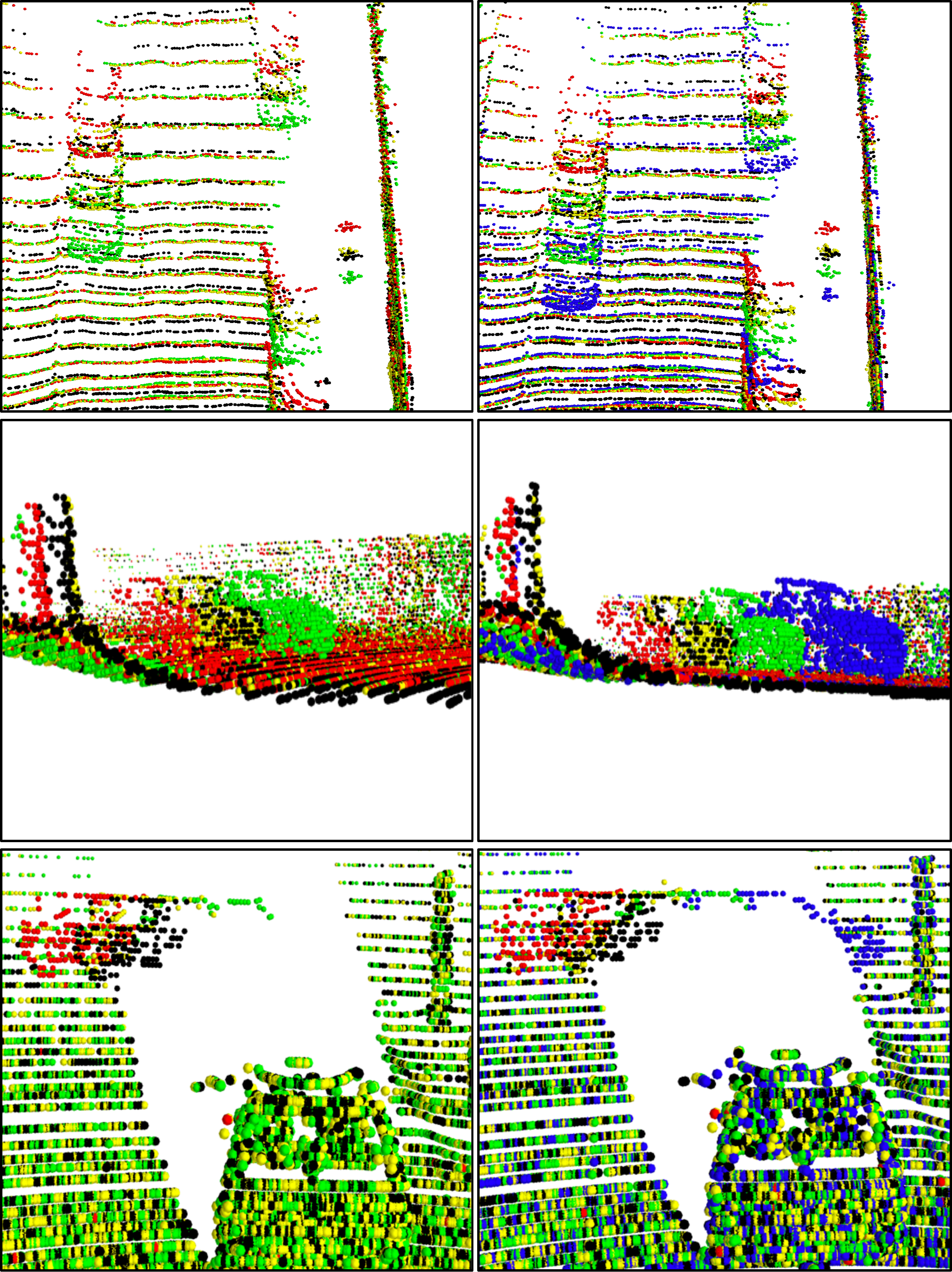}
    \caption{Visualization of SPCM-Net example results on the KITTI scene flow dataset with  three (left)  and  four (right)  preceding  frames as input. Different colors denotes points from a different frame number: \textcolor{red}{ frame $t-3$}, \textcolor{yellow}{frame $t-2$}, \textcolor{green}{frame $t-1$},  \textcolor{blue}{frame $t$}, and the \textbf{translated points} in black (frame $t-3$ + predicted scene flows). The translated points should highly overlap to points of frame $t-2$ for a good estimation. Left-column and right-column images show the results of using 4 and 3 input frames, respectively. Best viewed in color.}
    \label{fig:kitti_scene_flow}
\end{figure}

\textbf{Results}. We train and evaluate two model variants of SPCM-Net with three and four preceding frames as input to match MeteorNet's results. Increasing the number of frames from three to four achieves a consistent performance gain as expected. SPCM-Net significantly improves the results compared to the previous state-of-the-art method MeteorNet. Specifically, SPCM-Net achieves lower mean errors of $0.229$ and $0.194$ with three and four preceding frames, decreasing the errors of MeteorNet by $17.33\%$ and $22.71\%$, respectively.  The use of a recurrent cost volume to processes a point cloud sequence shows a better capability of extracting motion patterns for scene flow estimation than the 4D convolution approach of MeteorNet.
Fig. \ref{fig:kitti_scene_flow} visualizes some examples of the predicted scene flows.

\section{Related Work}\label{sec:related}
\subsection{Sequential Point Cloud Processing}

Our work is related to techniques for sequential point cloud processing. This area is fairly new but is a critical step towards general learning of scene dynamics.
To effectively process point cloud sequences, Fast and Furious (FaF) \citep{luo2018fast} proposes a network that jointly tackles 3D detection, tracking, and motion forecasting with a birds-eye view representation of point clouds and 3D convolutions. MinkowskiNet \citep{choy20194d} invents a generalized sparse tensor-based computing framework that allows handling point cloud sequences with 4D sparse convolutions. However, directly applying 4D convolutions on a voxelized sequence requires extensive computation. Furthermore, quantization errors during voxelization may cause performance drops when tackling problems requiring precise measurement, e.g., point-wise scene flow estimation.
Occupancy Flow \citep{niemeyer2019occupancy} learns a temporally
and spatially continuous vector field to perform the 4D reconstruction.
 More recently, PSTNet \citep{fan2021pstnet} designs a point spatiotemporal convolution to compute features from point cloud sequences with a hierarchical design. They apply their architecture to 3D action recognition. 
 3DV \citep{wang20203dv} proposes to use 3D dynamic voxels as the motion representation for depth videos and utilizes PointNet++ \citep{qi2017pointnet++}  for feature abstraction.  \cite{prantl2019tranquil} presented a new deep learning method that aims at capturing stable and temporally coherent features from point cloud sequences, via a novel temporal loss that extends the EMD loss to minimize the difference between estimated and ground-truth super-resolution point clouds in higher orders, i.e., positions, velocities, and accelerations. They introduced an additional \textit{mingling} loss term to push the individual points of a group apart, avoiding temporal mode collapse. Their method can be potentially adapted for our defined tasks as it could produce smooth motion for point cloud sequences. \cite{rempe2020caspr} made an important step to aggregate and
encode spatio-temporal changes of objects from point cloud sequences and learn Canonical Spatiotemporal Point Cloud Representation (CaSPR) via a Latent ODE approach. Their technique has been applied to various applications such as reconstruction, camera pose estimation, and correspondence estimation. P4Transformer \citep{Fan_2021_CVPR} introduced a new transformer-like network to model raw point cloud videos, where a novel point 4D convolution has been proposed to efficiently encode spatio-temporal local structures. P4Transformer shows superior performance on various benchmarks including 3D action
recognition \citep{li2010action,shahroudy2016ntu,liu2019ntu} and 4D semantic segmentation \citep{choy20194d}.

Recently, a collection of work has focused on learning scene dynamics from sequences of point clouds.
We consider MeteorNet \citep{liu2019meteornet} as a 4D extension of PointNet++ \citep{qi2017pointnet++} to handle temporal point cloud sequences by appending a 1D temporal coordinate to the 3D spatial coordinates. The resulting 4D coordinates help find spatiotemporal neighborhoods in 4D space. Both direct and chained-flow grouping are proposed to consider the spatiotemporal interaction of points such as the maximum travel distance and the motion direction of points. We show that SPCM-Net’s recurrent processing of sequences provides a stronger inductive bias compared to the 4D convolution of MeteorNet. MeteorNet is only evaluated on last-frame scene flow estimation at a relatively small scale, which is clearly distinct from the SSFE task proposed in this work. Their work also does not consider future prediction. PointRNN \citep{fan2019pointrnn} has been proposed to incorporate the \textit{flow embedding} layer of FlowNet3D \citep{liu2019flownet3d} into a recurrent unit for predicting future point clouds. We consider the \textit{flow embedding} layer as the \textit{point-to-set} matching cost, in contrast to SPCM-Net's recurrent cost volume that adopts a \textit{set-to-set} matching cost which is more robust to outliers (see Section \ref{sec:architectural}). PointRNN failed to adequately formalize the SPF task which limited their evaluation. They only focus on self-supervised training and evaluation, whereas we consider both supervised and self-supervised SPF in addition to supervised SSFE. 
Moreover, our experiments confirmed that SPCM-Net's recurrent cost volume for propagating point-wise spatiotemporal features across time is a more robust matching cost than PointRNN's flow embedding cost.
 Another self-supervised architecture, SPFNet, is proposed to tackle the SPF problem on self-driving datasets \citep{Weng2020_SPF2}. They investigated both point-based and range-map-based encoders to extract useful information from past frames, followed by LSTM-based decoders to predict future scene point clouds. The point-based encoder is similar to PointNet \citep{qi2017pointnet} to extract a global feature from each point cloud frame while the range-map-based encoder is only suitable for processing of LIDAR point clouds due to its range map representation. Their paper focused on trajectory forecasting and provided limited evaluation of the self-supervised SPF task. 
 
\subsection{Scene Flow Estimation}

 To estimate motion between frames, previous work \citep{dosovitskiy2015flownet,hui2018liteflownet,teed2020raft} tends to follow traditional optical flow approaches such as energy minimization \citep{horn1981determining} or warping-based methods \citep{brox2004high,bruhn2005lucas}. FlowNet is a generic deep learning architecture for optical flow estimation \citep{dosovitskiy2015flownet} that correlates feature vectors of image pairs at different
image locations. Since then, new work has been proposed to further improve its performance \citep{ilg2017flownet,ranjan2017optical,sun2018pwc}. 

Prior to the study of 3D scene flow estimation, flow estimation was concerned with motion across pairs of images. The three-dimensional scene flow is initially described in \citep{vedula1999three} as a 3D extension of 2D optical flow, where they formulate the estimation problem as a factor-graph-based energy minimization problem with hand-crafted SHOT descriptors \citep{tombari2010unique} for correspondence. Early work on scene flow estimation used multi-view geometry to associate salient image key points \citep{vedula1999three}. The problem has also been addressed by jointly optimizing registration and motion \citep{pons2007multi,huguet2007variational}.  

Recently, deep learning models have been proposed to estimate LiDAR flow with parametric continuous convolution layers \citep{wang2018deep}. The \textit{flow embedding} layer, introduced in FlowNet3D \citep{liu2019flownet3d}, encodes motion between two consecutive point clouds and has achieved competitive performance. HPLFlowNet \citep{gu2019hplflownet} instead estimates motion by converting point clouds into permutohedral lattices with bilateral convolutions to aggregate features. PointPWC-Net \citep{wu2019pointpwc} presents an end-to-end deep scene flow model to conduct scene flow estimation in a coarse-to-fine fashion.  FLOT \citep{puy20flot} finds the point correspondences between two points by adapting optimal transport with relaxed transport constraints to handle real-world imperfections. In \citep{mittal2020just}, the authors present a self-supervised approach based on nearest neighbors and cycle consistency with competitive results compared to supervised scene flow approaches. In \citep{pontes2020scene}, scene flow from point clouds is recovered and regularized with graph Laplacian \citep{bobenko2007discrete}. Our work draws upon model designs of the scene flow approaches while making further innovations to support recurrent processing of point cloud sequences to solve the SSFE and SPF tasks.

\subsection{Deep Learning on 3D  Point Clouds} 

Extensive research projects have been undertaken to develop modeling techniques aimed at automatically understanding 3D scenes and objects for numerous applications, such as 3D object classification \citep{klokov2017escape,qi2017pointnet,qi2017pointnet++,li2018so,dgcnn}, 3D object detection \citep{qi2018frustum,Shi_2019_CVPR}, 3D semantic labeling \citep{landrieu2018large,graham20183d,choy20194d,su2018splatnet}, and 3D instance segmentation \citep{pham2019jsis3d,yang2019learning,pan2020semanticposs}. Most prior work relies on transforming 3D data into regular representations such as voxels \citep{wu20153d} or 2D grids \citep{su2015multi} for processing. Here, context aggregation can be achieved easily with convolutions at relatively low resolutions due to the expensive computational overhead and memory footprint. To mitigate the issue, we have seen architectures such as OctNet  \citep{riegler2017octnet} and permutohedral lattice representations\citep{su2018splatnet} being proposed to achieve efficient memory allocation and computation without compromising resolution.

Recently, new work has emerged that directly processes raw and irregular point clouds \citep{qi2017pointnet,pham2019jsis3d,qi2017pointnet++,dgcnn} by applying MLPs in a point-wise fashion. To further capture local structures, follow-ups \citep{qi2017pointnet++,shen2018mining,dgcnn,thomas2019kpconv} have defined pseudo-convolutional operators where convolutions are instantiated as continuous kernels, assuming a continuous space for point clouds. However, this incurs an extra cost due to the use of greedy nearest neighbor search and point sampling algorithms for hierarchical processing. More recently, sparse tensor-based point cloud processing has been proposed to conduct sparse convolutions only on non-empty locations. Popular frameworks, such as SparseConvNet \citep{graham20183d}, MinkowskiEngine \citep{choy20194d}, and TorchSparse \citep{tang2020searching}, can conduct the sparse convolutions very efficiently based on their fast indexing structure.  

Up to now, the majority of this body of work is aimed at processing static point clouds. Less progress has been made on dynamic point cloud modeling, especially the motion estimation and future prediction of point cloud sequences, which is the focus of this paper.

\subsection{Self-supervised Learning}
Deep learning models have demonstrated the ability to obtain discriminative embeddings with unsupervised learning without providing any external supervision, i.e., supervision signals are generated from data itself \citep{lee2009convolutional,doersch2015unsupervised,srivastava2015unsupervised}. These representations could be used in downstream tasks or as strong initialization for supervised tasks. In point cloud processing, we have seen several works attempting to jointly learn multiple tasks including depth estimation, optical flow estimation, ego-motion estimation, and camera pose estimation based on 2D images \citep{yin2018geonet,zou2018df,lee2019cemnet}.  Other self-supervised point cloud tasks include point set generation \citep{fan2017point}, point cloud auto-encoder \citep{Yang_2018_CVPR}, point set registration \citep{fitzgibbon2003robust}. We refer the interested reader to \citep{guo2019deep} for a comprehensive examination. In this paper, we have utilized geometric losses, i.e., chamfer distance and earth mover’s distance \citep{rubner2000earth}, for self-supervised learning of future prediction from point cloud sequences. Advanced techniques such as Laplacian regularization or local smoothness \citep{wu2019pointpwc,pontes2020scene} could be further utilized to regularize the learning and improve the performance.

\subsection{Spatiotemporal Learning}
RNNs and their variants are widely used in sequence prediction while having difficulty in applying directly to structured data such as videos due to the ignorance of handling the spatial arrangement of data. To address this, the convolutional LSTM (ConvLSTM) \citep{xingjian2015convolutional} is proposed to capture local spatial correlations and replace the fully connected layer in the recurrent state transition with a convolution operation. The spatiotemporal LSTM \citep{wang2017predrnn} further extends ConvLSTM by introducing a novel recurrent unit that can deliver memory states both vertically (across recurrent layers) and horizontally (across time). Numerous follow-up works have been proposed along this direction \citep{wang2018predrnn++,wang2019eidetic}. In CubicLSTM \citep{fan19cubiclstm}, the authors extend the ConvLSTM by utilizing two states (the temporal state and the spatial state) with independent convolutions. All of these LSTMs can be applied to spatiotemporal data. However,  additional designs (see Section \ref{sec:method}) are required to apply them to 3D point cloud sequences because the point cloud sequences are unstructured and orderless. Our work demonstrates one way in which these methods can be applied to point cloud sequences.

\section{Discussion}\label{sec:discussion}

 Our experimental results showed that SPCM-Net achieves superior performance compared to state-of-the-art SFE models on the SSFE task by leveraging temporal coherence of points over many frames.
 We attribute this to the recurrent cost volume layer, which effectively propagates point-wise spatiotemporal information across time.
 Empirically, we observed a large reduction in outlier predictions which helped improve the overall scene flow estimation performance.
SPCM-Net also produces competitive SPF results under supervised and self-supervised settings compared to the best prior model.
We achieved state-of-the-art SPF performance by first pre-training on the SSFE task before fine-tuning on SPF. 

This evaluation is conducted on a newly introduced benchmark for SSFE and SPF.
As shown by our qualitative results, the ground truth supervision provided for the two synthetic datasets SFT3D and VKS enables a rigorous and principled comparison between competing models.
Both datasets offer unique challenges for future study; in particular, the VKS dataset contains multiple dynamic objects in each frame and requires learning physical properties of vehicles for accurate estimation and prediction.
The real-world SAG dataset also contains its own set of challenges, particularly related to handling outliers during training.
We expect this benchmark to be pivotal for standardizing training and evaluation protocols for future work on SSFE and SPF.

\subsection{Limitations and Future Work}
In this work, our studied tasks focus on relatively low-level dynamic point cloud processing that aims to predict the point-wise motion of sequences. We expect that object category information and motion smoothness priors could help improve performance. This, however, requires defining novel tasks and proposing new benchmarks. Also, the self-supervised Chamfer distance objective used in our paper is designed for point matching between two point sets at each timestep. Objective functions that take into account temporal correspondence \emph{across} frames, like those used for multi-frame data association in multi-object tracking~\citep{emami2020machine}, are needed to provide stronger supervision signals.

We can suggest further promising directions for future research.
First, better modeling of occlusion is needed to further improve the scene flow estimation, e.g., \citep{ouyang2020occlusion}. 
Second, the self-supervised SPF remains an open problem.
Currently, models make an assumption that points in the current predicted future frame are translated from points in the previous predicted frame with a motion offset.
In real-world applications, due to the nature of how each point cloud is generated, this assumption does not hold.
This leads to isolated points being improperly matched, which then leads to noisy training signals.
A promising alternative to nearest-neighbor losses like Chamfer distance is adversarial learning \citep{zuanazzi2020adversarial} where point
clouds are mapped to a latent space in which a robust distance metric can be computed.
\section{Conclusion}\label{sec:conclusion}

In this paper, we introduced \emph{sequential scene flow estimation} (SSFE) for point cloud sequences, which is a novel extension of the well-studied scene flow estimation task to multiple frames.
We proposed the SPCM-Net architecture to solve this task as well as the related sequential point cloud forecasting (SPF) task.
To help advance future research, we collected and presented a new benchmark consisting of three point cloud sequence datasets containing diverse backgrounds and multiple object motions in synthetic and realistic environments.
Our benchmark uniquely contains ground truth annotations for multi-step scene flow which current SFE datasets lack, which should be pivotal to future research.

\section*{Acknowledgements} This work is supported by NSF CNS 1922782, by the Florida Dept. of Transportation (FDOT) and FDOT District 5. The opinions, findings and conclusions expressed in this publication are those of the author(s) and not necessarily those of the Florida Department of Transportation or the National Science Foundation. 

\section*{Declarations}
\noindent \textbf{Conflict of interest} The authors declare that they have no conflict of
interest.


%
%

\bibliographystyle{spbasic}

\bibliography{ref}   

\appendix

\section{Appendix}





\subsection{Additional Evaluations}

\begin{table*}[hbt!]
    \centering
\caption{Evaluation results on VKS for the SSFE task with the goal to test generalization to unseen driving scenarios.}\label{tab:vks_split}
    \resizebox{\textwidth}{!}{
    \begin{tabular}{l|c|cccc|cc}       
         \hline\noalign{\smallskip}
        \multicolumn{8}{c}{\textbf{VKS Train Split}}
         \\ 
         \hline\noalign{\smallskip}
        Method         & EPE3D$\downarrow$           & Acc3DS$\uparrow$          & Acc3DR$\uparrow$          & Outliers3D$\downarrow$  & RectOutliers3D$\downarrow$ & EPE2D$\downarrow$           & Acc2D$\uparrow$      \\
        \noalign{\smallskip}\hline\hline
 FlowNet3D \citep{liu2019flownet3d} &  0.0422      & 0.8790      & \textbf{0.9718}   & 0.2043      & 0.0855 & 3.9368   & 0.9159     \\
        FLOT~\citep{puy20flot}  & \textbf{0.0396}  & 0.8625  & 0.9316 & 0.2174  & 0.1118 & 2.5706  & 0.8939 \\
        PointPWC-Net \citep{wu2019pointpwc} & 0.0588   & 0.9202   &  0.9542   & \textbf{0.1960} &  \textbf{0.0824} & 2.4082  & \textbf{0.9306}  \\
        \hline 
        \hline 
         SPCM-Net (Ours) & 0.0618   &  \textbf{0.9212}   &  0.9556   & 0.2061  &  0.0877 & \textbf{2.4044}  &  0.9180  \\
         \hline 
         \hline\noalign{\smallskip}
        \multicolumn{8}{c}{\textbf{VKS Test Split}}
         \\ 
         \hline\noalign{\smallskip}
        Method         & EPE3D$\downarrow$           & Acc3DS$\uparrow$          & Acc3DR$\uparrow$          & Outliers3D$\downarrow$  & RectOutliers3D$\downarrow$ & EPE2D$\downarrow$           & Acc2D$\uparrow$      \\
        \noalign{\smallskip}\hline\hline
 FlowNet3D \citep{liu2019flownet3d} &  0.0550     & 0.6877     & \textbf{0.9123}  & 0.3974      & 0.2888 & 2.3364   & 0.8368    \\
        FLOT~\citep{puy20flot}  & 0.0689  & 0.7471   & 0.8708 & 0.3484  & 0.2506 & 3.4343  & 0.8360 \\
        PointPWC-Net \citep{wu2019pointpwc} & 0.0478   & 0.7972   &  0.8993   & 0.3458 &  0.2440 & 2.2210  & 0.8556  \\
        \hline 
        \hline 
         SPCM-Net (Ours) & \textbf{0.0477}   &  \textbf{0.8106}   &  0.9082   & \textbf{0.3298}  &  \textbf{0.2268} & \textbf{2.0846}  &  \textbf{0.8678}  \\
         \hline 
         
    \end{tabular}
    }
\end{table*}
\begin{table}[hbt!]
\centering
\caption{Evaluation results on nuScenes for the SPF task.}
\normalsize
 \resizebox{0.5\textwidth}{!}{  
\begin{tabular}{l|c|c|c}
\hline \noalign{\smallskip}
 \multirow{2}{*}{Method}   & \multicolumn{3}{c}{\textbf{NuScenes}}       \\\noalign{\smallskip} \cline{2-4} \noalign{\smallskip}
                       & CD$\downarrow$          & EMD$\downarrow$ & SD$\downarrow$    \\ \noalign{\smallskip} \hline \hline
 PointNet++~\citep{qi2017pointnet++} + LSTM  & \textbf{0.6176} & 0.8334 & \textbf{0.6920} \\
 PointRNN~\citep{fan2019pointrnn}& 0.9750 & 0.9878 & 1.0969 \\  \hline \hline
SPCM-Net (Ours) & 0.6339 & \textbf{0.7858} & 0.7113 \\ \hline
\end{tabular}
}
\label{tab:nuscenes_future_split}\label{sec:spf_vks_split}
\end{table}
This section provides additional results supporting the evaluations presented in the Experiments.

\subsubsection{Additional SPF results on nuScenes} \label{sec:spf_nuscenes_future}

The nuScenes dataset \citep{nuscenes2019} is a large-scale public autonomous driving dataset, which contains $850$ publicly available scenes in total collected in both Boston and Singapore, which are known for dense traffic and highly challenging driving situations. 15h of driving data (242km traveled at an average of 16km/h) was collected with the dataset containing 68 driving logs for training and 15 driving logs for testing. The LiDAR data was captured by a Velodyne 32-beam LiDAR.

The nuScenes dataset does not provide ground-truth annotations for scene flow. Therefore, we adopt metrics without requiring any annotation. Given the fact that forecasting future point clouds on real-world datasets is challenging due to rapid changes in the vehicle's surroundings, we focus on short-term prediction for the nuScenes dataset. With each driving log providing a point cloud sequence, we repeatedly sample from it by randomly choosing 10 successive point clouds for training and testing (consecutively sampling every other frame and repeating 10 times). To achieve reasonable computation times while staying within memory limits, we sample a fixed number of points for each frame and remove the ground points to reduce the bias caused by the flattened geometry of the ground. Specifically, $2,048$ points are randomly sampled from every point cloud frame in the sequence. Following the practice of \citep{Wu_2020_CVPR}, the point clouds are cropped to extract the region defined by  $[-32, 32] \times [-8, 8] \times [-1.3, 2]$ meters, which corresponds to the XYZ range.

The experimental results of nuScenes are summarized in Table~\ref{tab:nuscenes_future_split}. Our SPCM-Net achieves a competitive result compared to other baselines in all metrics. One interesting observation is that both SPCM-Net and  PointNet++~\citep{qi2017pointnet++} + LSTM outperform PointRNN \citep{fan2019pointrnn}
significantly. We suspect that static points dominate on the nuScenes dataset such that most of the points only contain ego-motion. Therefore, it prefers models with a better capability of extracting global information. 

\subsubsection{Additional SSFE results on VKS}\label{sec:ssfe_vks_split}

\begin{table}[hbt!]
\centering
\caption{Evaluation results on VKS for the SPF task with the goal to test generalization to unseen driving scenarios.}
\normalsize
 \resizebox{0.5\textwidth}{!}{ 
\begin{tabular}{l|c|c|c|c}
\hline \noalign{\smallskip}
 \multirow{2}{*}{Method}  & \multicolumn{4}{c}{\textbf{VKS Train Split}}      \\\noalign{\smallskip} \cline{2-5} \noalign{\smallskip}
                       & ADE $\downarrow$           &  FDE$\downarrow$          & CD$\downarrow$          & EMD$\downarrow$    \\ \noalign{\smallskip} \hline \hline
 PointNet++~\citep{qi2017pointnet++} + LSTM  & 0.9971  & 1.6726 & 0.5783 & 1.2601  \\
 PointRNN~\citep{fan2019pointrnn}& \textbf{0.2201}  & \textbf{0.3292}  & \textbf{0.1242} & \textbf{0.2856}   \\   \hline \hline
SPCM-Net (Ours) & 0.2535  & 0.4327  & 0.1418  & 0.3315   \\ \hline
\hline \noalign{\smallskip}
 \multirow{2}{*}{Method}  & \multicolumn{4}{c}{\textbf{VKS Test Split}}      \\\noalign{\smallskip} \cline{2-5} \noalign{\smallskip}
                       & ADE $\downarrow$           &  FDE$\downarrow$          & CD$\downarrow$          & EMD$\downarrow$    \\ \noalign{\smallskip} \hline \hline
 PointNet++~\citep{qi2017pointnet++} + LSTM  & 1.4951 & 2.4608 & 0.8209 & 1.8308  \\
 PointRNN~\citep{fan2019pointrnn}& 0.3334  & 0.5562  & 0.1798 & 0.4095   \\   \hline \hline
SPCM-Net (Ours) & \textbf{0.3291}  & 0.5853  & 0.1824  & \textbf{0.4068}  \\ \hline
\end{tabular}
}

\label{tab:vks_future_split}
\end{table}
In Section \ref{sec:vks}, we have evaluated the performance of VKS, providing a preliminary prototype evaluation for traffic scenes. Table \ref{tab:vks_split} further supplements it with the evaluation results on the VKS dataset with a new split to report the generalization capability of unseen driving scenarios. Recall from Section \ref{sec:vks_dataset} that the original virtual KITTI contains five scenes of crowded
urban area (Scene01), busy intersections (Scene02, Scene06), long road in the forest (Scene18), and highway driving scene (Scene20), and we initially have chosen the train and test splits both containing examples of all scenes. We made a further exploration to create the train and test splits such that they do not have any overlap on scenarios. To do so, we held out the highway driving scene (Scene20) as the test split and use other scenes as the train split. This leads a total of $1,876$ train and $1,474$ test point cloud sequences. We didn't try other possible split combinations as it would end up with excessive number of experiments to run. Instead, we will provide the possibility to try other combinations in our implementation for future research.

As expected, the SSFE performance of all fully supervised methods drops slightly when moving from the training scenes to the unseen scenes (Table \ref{tab:vks_split}). Remarkably though, obtaining comparable or worse results on the training split, our SPCM-Net outperforms other methods on the unseen scenes, implying that it has a better generalization capability on the SSFE task.

\subsubsection{Additional SPF results on VKS}

Similarly, we also evaluate the prediction performance on VKS following the same split as done in previous Section \ref{sec:ssfe_vks_split}. Our SPCM-Net still achieves a comparable performance compared to PointRNN \citep{fan2019pointrnn}, drawing a similar conclusion as in Section \ref{sec:vks}.

\clearpage
\newpage
\end{document}